\documentclass[10pt,twocolumn,letterpaper]{article}

\usepackage{cvpr}

\usepackage[utf8]{inputenc} %
\usepackage[T1]{fontenc}    %
\usepackage{url}            %
\usepackage{booktabs}       %
\usepackage{amsfonts}       %
\usepackage{nicefrac}       %
\usepackage{microtype}      %

\usepackage{enumitem}
\usepackage{algpseudocode}
\usepackage{xcolor}
\usepackage{array}
\usepackage{multirow}
\usepackage{algorithm}
\usepackage{subcaption}
\usepackage[normalem]{ulem}
\usepackage{xparse}
\usepackage{pifont}
\usepackage{bm}
\usepackage{threeparttable}
\usepackage{etoolbox}
\usepackage{listings}
\usepackage{mwe} %
\usepackage{makecell}
\usepackage{color, colortbl}
\usepackage{tabularx}
\usepackage[accsupp]{axessibility}
\usepackage{soul}
\usepackage{tabu}

\usepackage[scaled=0.85]{DejaVuSansMono}

\usepackage{tikz}
\usepackage{pgfplots}
\pgfplotsset{compat=1.16}
\usepgfplotslibrary{groupplots}
\usetikzlibrary{patterns}

\definecolor{citecolor}{HTML}{0071bc}
\usepackage[breaklinks=true,bookmarks=false,colorlinks,bookmarks=false, citecolor=citecolor]{hyperref}

\usepackage[capitalize]{cleveref}
\crefname{section}{\S}{\S\S}
\crefname{subsection}{\S}{\S\S}
\crefformat{table}{Table~#2#1#3}
\crefformat{figure}{Figure~#2#1#3}
\crefformat{equation}{Eq~#2#1#3}

\makeatletter
\DeclareRobustCommand\onedot{\futurelet\@let@token\@onedot}
\def\@onedot{\ifx\@let@token.\else.\null\fi\xspace}

\def\eg{\emph{e.g}\onedot} 
\def\ie{\emph{i.e}\onedot} 
\def\cf{\emph{cf}\onedot} 
\def\etc{\emph{etc}\onedot}

\makeatother

\newlength\savewidth

\newlength\thinwidth

\definecolor{Gray}{gray}{0.92}
\definecolor{DarkGray}{gray}{0.5}
\newcolumntype{a}{>{\columncolor{Gray}}c}
\newcolumntype{H}{>{\setbox0=\hbox\bgroup}c<{\egroup}@{}}
\definecolor{LightCyan}{rgb}{0.88,1,1}
\definecolor{altRowColor}{gray}{0.92}
\definecolor{highlightRowColor}{gray}{0.9}
\newcommand{\colorrow}{\rowcolor{highlightRowColor}}
\DeclareRobustCommand{\colorrowtext}[0]{{\sethlcolor{highlightRowColor}\hl{gray}}}
\soulregister{\colorrowtext}{1}

\definecolor{amber}{rgb}{1.0, 0.75, 0.0}
\definecolor{canaryyellow}{rgb}{1.0, 0.94, 0.0}
\definecolor{Black}{rgb}{0.0, 0.0, 0.0}

\definecolor{GrayNumber}{gray}{0.5}
\definecolor{GrayXMark}{gray}{0.7}
\newcommand{\xmark}{ {\color{GrayXMark} \ding{55}} } %

\definecolor{ImageDark}{rgb}{0,0.3,0.8}
\definecolor{VideoDark}{rgb}{.5,.0,.5}
\definecolor{ThreeDDark}{rgb}{0,.5,0}
\colorlet{Image}{ImageDark!20!white}
\colorlet{Video}{VideoDark!20!white}
\colorlet{ThreeD}{ThreeDDark!20!white}
\colorlet{ImageLight}{ImageDark!70!white}
\colorlet{VideoLight}{VideoDark!70!white}
\colorlet{ThreeDLight}{ThreeDDark!70!white}
\newcommand{\VideoMark}[0]{*}
\newcommand{\ImageMark}[0]{square*}

\newcolumntype{i}{>{\columncolor{Image}}c}
\newcolumntype{v}{>{\columncolor{Video}}c}
\newcolumntype{t}{>{\columncolor{ThreeD}}c}
\newcolumntype{I}{>{\columncolor{ImageLight}}c}
\newcolumntype{V}{>{\columncolor{VideoLight}}c}
\newcolumntype{T}{>{\columncolor{ThreeDLight}}c}

\newcommand{\OURS}{OmniMAE\xspace}

\newcommand{\swin}{Swin\xspace}
\newcommand{\swinB}{Swin-B\xspace}

\newcommand{\vit}{ViT\xspace}
\newcommand{\vitB}{ViT-B\xspace}
\newcommand{\vitL}{ViT-L\xspace}
\newcommand{\vitH}{ViT-H\xspace}

\newcommand{\imnet}{ImageNet\xspace}
\newcommand{\imnetShort}{IN1K\xspace}

\newcommand{\kinetics}{Kinetics-400\xspace}
\newcommand{\kineticsShort}{K400\xspace}

\newcommand{\placesThree}{Places-365\xspace}
\newcommand{\placesThreeShort}{P365\xspace}

\newcommand{\inat}{iNaturalist-2018\xspace}
\newcommand{\inatShort}{iNat18\xspace}

\newcommand{\sthsth}{Something Something-v2\xspace}
\newcommand{\sthsthShort}{SSv2\xspace}
\newcommand{\epic}{EPIC-Kitchens-100\xspace}
\newcommand{\epicShort}{EK100\xspace}

\newcommand{\app}{\raise.17ex\hbox{$\scriptstyle\sim$}}

\newcommand{\totalPatch}{N}
\newcommand{\totalMaskPatch}{M}

\newcommand{\sota}{state-of-the-art\xspace}

\newcommand{\autoencoders}{autoencoders\xspace}

\newcommand{\vmaeFull}{SpatioTemporal-MAE\xspace}
\newcommand{\vmae}{ST-MAE\xspace}
\newcommand{\videomae}{\vmae}
\newcommand{\dino}{DINO\xspace}
\newcommand{\ibot}{iBOT\xspace}
\newcommand{\mae}{MAE\xspace}
\newcommand{\maskedFeatV}{MaskedFeat\xspace}

\def\confName{CVPR}
\def\confYear{2023}

\begin{document}

\title{\OURS: Single Model Masked Pretraining on Images and Videos}

\author{
  Rohit Girdhar$^{*}$ \quad
  Alaaeldin El-Nouby$^{*}$ \quad
  Mannat Singh$^{*}$ \\
  Kalyan Vasudev Alwala$^{*}$ \quad
  Armand Joulin \quad
  Ishan Misra$^{*}$ \\
  FAIR, Meta AI \\
  {\footnotesize \url{https://github.com/facebookresearch/omnivore}}
}

\twocolumn[{%
\renewcommand\twocolumn[1][]{#1}%
\maketitle
\begin{center}
    \captionsetup{type=figure}
    \includegraphics[width=0.9\linewidth]{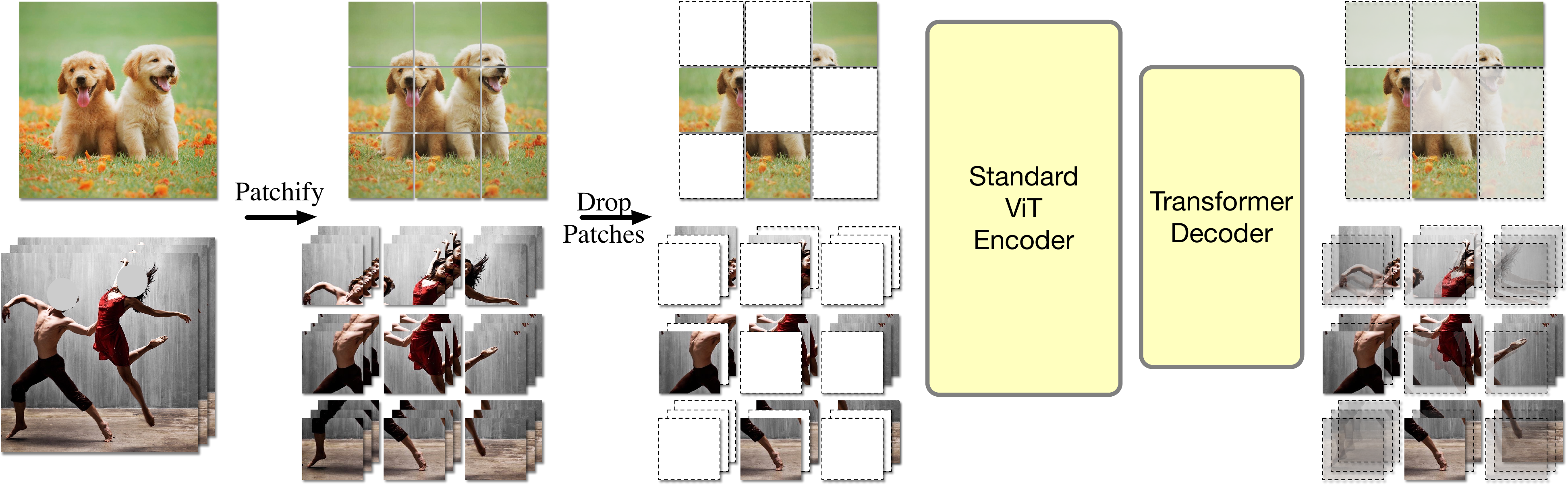}
    \captionof{figure}{{\bf \OURS} is a single model for images and videos that is trained using masked
    autoencoding~\cite{he2021masked}.
    We use a plain Vision Transformer~\cite{dosovitskiy2020image} architecture but with spatio-temporal patches as input.
    At training, we `patchify' the visual input (images or videos), and feed the encoder only a
    subset of the patches. The decoder reconstructs the pixels for the missing patches using the
    encoder's output. %
    The encoder-decoder
    model is trained using a pixel reconstruction loss. After training, our single plain Transformer
    encoder performs competitively compared to specialized architectures on downstream image and
    video recognition tasks. }
    \label{fig:approach}
\end{center}%
}]

\makeatletter{\renewcommand*{\@makefnmark}{}
\footnotetext{$^*$Equal technical contribution.}\makeatother}

\begin{abstract}
    Transformer-based architectures have become competitive across a variety of visual domains, most
    notably images and videos. While prior work studies these modalities in isolation, having a
    common architecture suggests that one can train a single unified model for multiple visual
    modalities. Prior attempts at unified modeling typically use architectures tailored for vision
    tasks, or obtain worse performance compared to single modality models. In this work, we show
    that masked autoencoding can be used to train a simple Vision Transformer on images and videos,
    without requiring any labeled data. This single model learns visual representations that are
    comparable to or better than single-modality representations on both image and video benchmarks,
    while using a much simpler architecture.
    Furthermore, this model can be learned by dropping 90\% of the image and 95\% of the video
    patches, enabling extremely fast training of huge model architectures.
    In particular, we show that our single ViT-Huge %
    model can be
    finetuned to achieve 86.6\% on \imnet and 75.5\% on the challenging \sthsth video benchmark, setting a new \sota.

\end{abstract}

\section{Introduction}
\label{sec:intro}

The Transformer architecture~\cite{vaswani2017attention} is rapidly becoming competitive across the
different visual modalities in Computer Vision, from
images~\cite{dosovitskiy2020image,liu2021swin,fan2021multiscale,touvron2021training}, to
3D~\cite{misra2021end,zhao2020point,liu2021group} and
videos~\cite{liu2021video,fan2021multiscale,bertasius2021is,arnab2021vivit,girdhar2021anticipative,girdhar2019video}. This convergence toward
a unified architecture naturally suggests that we should be able to train a single model that works
across different visual modalities.
However, recent attempts to train unified models either lead to worse performance compared to single modality
models~\cite{anonymous2021polyvit}, or require the use of an alternative
architecture~\cite{girdhar2022omnivore}, namely the Swin Transformer~\cite{liu2021swin}, with
inductive biases tailored towards vision tasks. While specialized Transformer architectures for
vision~\cite{liu2021swin,liu2021video,fan2021multiscale,wang2021pyramid} can offer better
performance for visual modalities, they lose the generality and flexibility of the vanilla
Transformer, making it harder to later model different domains like text, speech, 3D \etc in
multi-modal architectures.

In this work, we train a single vanilla Transformer that works for both images and videos, as illustrated in~\cref{fig:approach}.
To this end, we leverage the findings of several recent works on the use of the masked pretraining~\cite{devlin2018bert}
to greatly improve the training and performance of Transformers in the domain of
images~\cite{bao2021beit,he2021masked,wang2021bevt,xie2021simmim},
videos~\cite{tong2022videomae,wei2021masked,wang2021bevt} or across text, audio and
images~\cite{baevski2022data2vec}.
We show that this masked pretraining is a viable strategy to pretrain a unified `omnivorous' Transformer across visual modalities.
In particular, we consider the Masked Auto-Encoding ({\bf MAE}) approach~\cite{he2021masked} to train an {\bf
Omni}vorous visual encoder~\cite{girdhar2022omnivore}. The resulting {\bf \OURS} model learns from all the modalities with the
same objective function and does not require any supervision.

Using a masked pretraining objective has several advantages over
supervised objectives~\cite{girdhar2022omnivore,anonymous2021polyvit} or discriminative
self-supervised objectives~\cite{he2020momentum,chen2020simple,caron2021emerging}. First, as opposed to supervised
objectives, a general-purpose unsupervised loss does not require any human labeling effort. As a
result, it is robust to biases introduced by a %
predefined set of labels~\cite{goyal2022vision}. Moreover, it does not require a multi-head %
architecture to incorporate supervision from each of the label spaces corresponding to each
modality, which is hard to maintain and scale with new %
modalities. Second, although discriminative self-supervised methods produce superior frozen
features compared to reconstruction objectives, they are non trivial to scale in model and data
size~\cite{goyal2019scaling}. Our masked pretraining objective is simple, efficient to train, and
scales to different visual modalities with minimal changes.

\par \noindent \textbf{Contributions.}
\textbf{(1)} We show that the simple Vision Transformer architecture (\vit)~\cite{dosovitskiy2020image} originally designed for images can naturally be applied on videos, and videos and images jointly.
\OURS is a single \vit-based model for videos and images that outperforms architectures and models specifically designed for either modality.
\textbf{(2)} Prior and concurrent work design self-supervised methods and architectures for either image or video and we find that these models do not transfer well across modalities.
\OURS is the first single self-supervised model that achieves good performance on both modalities.
\textbf{(3)} We show that our joint training using both images and videos enables us to use much higher masking ratios than any prior work for training MAE.
Since \vit can processes only the non-masked input, we train \OURS models with only 10\% of image and 5\% of video patches.
This enables us to train large (650M parameter) models with a $\sim7\times$ and $\sim11\times$ reduction in compute and memory on images and videos.
\textbf{(4)} Finally, we propose improvements to the MAE training.
We show that repeating samples in a mini-batch reduces dataloading (and thus training) time without loss in final transfer performance.
Sample replication is particularly useful for masked pretraining as the unmasked patches are different across sample replicas.
We also show that using a shallow shared decoder for videos and images leads to better performance while reducing the number of parameters by $2-4\times$.

\section{Related Work}
\label{sec:related}

Our work builds upon research in self-supervised learning, masked pretraining and unified modeling
in computer vision.

\par \noindent \textbf{Self-supervised learning.} In recent years, self-supervised approaches have
been dominated by joint embedding methods which can rely on different objectives including
contrastive~\cite{hadsell2006dimensionality,oord2018representation,chen2020simple,he2020momentum,misra2020self,wu2018unsupervised,junnan2021prototypical,tian2019contrastive},
non-contrastive~\cite{chen2020exploring,ermolov2020whitening,zbontar2021barlow,bardes2022vicreg},
clustering~\cite{asano2019self,caron2018deep,caron2020unsupervised,yan2020cluster} or
self-distillation~\cite{caron2021emerging,grill2020bootstrap,zhou2021ibot,baevski2022data2vec}. Such
methods are trained to learn invariance to a set of pre-defined transformations which results in
image descriptors with a strong
linear probing and KNN performance. However, such methods can be challenging to scale since they can
suffer from instabilities~\cite{chen2021empirical}. Additionally, the strongest performance is
typically obtained with the help of augmentations like multi-crop \cite{caron2020unsupervised,
caron2021emerging, zhou2021ibot} which can be hard to apply at scale due their compute and memory
overhead.

\par \noindent \textbf{Masked pretraining.} We build upon masked prediction methods where the
representation is learned by predicting masked parts of the input. Such methods have recently gained
popularity given their immense success in NLP. In particular, BERT~\cite{devlin2018bert} showed that
masked language modeling by predicting a subset of masked words is a powerful pre-training objective
and leads to impressive finetuning performance on various downstream tasks. In computer vision,
input reconstruction methods have a rich history with non-linear PCA~\cite{kramer1991nonlinear},
sparse reconstruction~\cite{olshausen1996},
\autoencoders~\cite{bourlard1988auto,hinton1993autoencoders,gallinari1987memoires,le1987modeles},
RBMs~\cite{salakhutdinov2009deep} \etc.  Masked prediction methods can be viewed as a special case
of denoising \autoencoders~\cite{gallinari1987memoires,vincent2008extracting} where the input
`noise' is a masking function. An example of such a method that uses masking as noise is context
encoders~\cite{pathak2016context}. With the recent and rapid rise of Vision
Transformers~\cite{dosovitskiy2020image}, masked prediction was revisited by multiple efforts.
SiT~\cite{atito2021sit} replaces variable sized patches in the image with random noise and trains a
\vit model for reconstruction, among other objectives. BEiT~\cite{bao2021beit} moves a step closer
to BERT with replacing full patches with mask tokens and training a \vit encoder to predict the
discrete visual words of masked patches using a cross-entropy loss. Masked prediction has also shown
impressive performance for specialized vision transformer architectures such as
MViT~\cite{li2022improved,fan2021multiscale,wei2021masked} and \swin
transformer~\cite{liu2021swin,liu2022swinv2,xie2021simmim}.
SimMIM~\cite{xie2021simmim} and MaskedFeat~\cite{wei2021masked} predict pixel values and HOG features of the masked patches using a
Swin-v2~\cite{liu2022swinv2} and MViT-v2~\cite{li2022improved} backbones respectively.
Finally, SplitMask~\cite{el2021large} studied the interesting properties of masked prediction
methods in terms of high sample efficiency.

Of particular interest to \OURS, masked autoencoders (MAE)~\cite{he2021masked} demonstrated
impressive scaling properties by utilizing a patch dropping strategy for masked patches accompanied
with a high masking ratio of 75\%. Under this setting, the encoder only process a small subset of
the image patches followed by a relatively low capacity decoder which reconstructs the image pixels.
This property is even more crucial for video representation learning given the large number of
patches, and a concurrent work,
\videomae~\cite{tong2022videomae}, shows that \mae pretraining with an even higher masking ratio of
90\% works well and obtains strong finetuning performance on downstream video recognition
benchmarks.
Notably, the efficiency gained by patch dropping is specific to vanilla ViTs, and multiscale
architectures such as MViT and \swin are unable to benefit due to their design. Hence, we use the
simple \vit as our architecture and show that it can be trained efficiently and jointly for images
and videos using extremely high masking ratios (90-95\% on both modalities), and yet perform
competitively to specialized architectures like MViT~\cite{fan2021multiscale} and \swin~\cite{liu2021swin}.

\par \noindent \textbf{Unified modeling and multi-modal learning.} Multi-modal learning in computer
vision  has a long history that includes training using images and text
\cite{lu202012,castrejon2016learning,miech2020end,karpathy2015deep,gong2014improving}, video and
optical flow \cite{simonyan2014two,feichtenhofer2016convolutional}, and video and audio
\cite{morgado2021audio,morgado2021robust,owens,arandjelovic2017look}. The majority of such methods
rely on training a separate backbone for each modality as well as the availability of alignment
across modalities. More recently, Omnivore~\cite{girdhar2022omnivore} was proposed for joint
training of multiple modalities like images, videos and single-view 3D, attaining a strong
performance in each of the modality specific benchmarks with a single shared trunk.
PolyViT~\cite{anonymous2021polyvit} co-trains a shared transformer encoder using images, videos and
audio data and provides a competitive performance on various downstream tasks for each modality. The
aforementioned methods differ compared to \OURS in that they are trained with supervised learning and
require human annotations. BEVT~\cite{wang2021bevt} tackles BERT pre-training for videos and
proposes that jointly pre-training using static images improves the finetuning performance of video
recognition benchmarks. Unlike \OURS, BEVT uses the specialized \swin transformer architecture with
separate decoder heads for images and videos. Thus, BEVT cannot drop patches while training which
can limit its scalability. Furthermore, it relies on a tokenizer which must be trained apriori, and
the tokenizer training itself can affect the model's performance.

\section{\OURS}
\label{sec:approach}

Our goal is to pretrain a single unified model for images and videos. Rather than use specialized
architectures tailored for a visual modality, we build upon the vanilla Vision
Transformer (ViT)~\cite{dosovitskiy2020image} architecture that has limited inductive biases for
vision. For pretraining, we extend the simple self-supervised masked auto-encoding (MAE)
approach~\cite{he2021masked}. The original architecture and pretraining method are tested only on
images, and we show simple design decisions for a unified model.

\subsection{Training \OURS jointly on images and videos}

We illustrate our method in~\cref{fig:approach}.
For pretraining, we use an encoder-decoder architecture where the encoder only operates on a
`non-masked' subset of the input.
The decoder predicts the pixel values for the entire input, \ie, masked and non-masked pixels. The
model is trained to minimize the reconstruction error for the masked (unseen) part of the input.
After pretraining, we evaluate the encoder by transfer learning (the decoder is discarded). Next, we
describe the pretraining details.

\textbf{Images and videos as spatio-temporal patches.}
The input image or video can be represented as a 4D tensor of shape $T \!\times\! H\! \times\! W\! \times 3$
where $T$ is the temporal dimension and $H, W$ are the spatial dimensions, and $3$ represents the
color channels. We treat images as being single-frame videos with $T=1$. The input is split into
$\totalPatch$ spatio-temporal patches, each of size $t\!\times\! h\!\times\! w\! \times\!3$~\cite{girdhar2022omnivore}.

\textbf{Omnivorous visual encoder.}
We use an omnivorous~\cite{girdhar2022omnivore} visual encoder that processes both images and video
using the same parameters.
The encoder operates on the $\totalPatch$ spatio-temporal patches from the images and videos. The
encoder can naturally handle variable number $\totalPatch$ of patches from images and videos as it
uses the Transformer architecture~\cite{vaswani2017attention}. The encoder shares the same
parameters for both image and video inputs and processes them in the same way to output per-patch
embeddings.

\begin{figure*}[t]
    \centering
    \vspace{0pt}  %
    \resizebox{\linewidth}{!}{
        \begin{tikzpicture}
    \pgfmathsetmacro{\NPlotsFirstAx}{5}
    \pgfmathsetmacro{\NPlotsSecondAx}{5}
    \pgfmathsetmacro{\SumPlots}{\NPlotsFirstAx+\NPlotsSecondAx}
    \pgfmathsetmacro{\WdFirst}{\NPlotsFirstAx/\SumPlots}
    \pgfmathsetmacro{\WdSecond}{\NPlotsSecondAx/\SumPlots}
    \begin{groupplot}[
        group style={
            group name=plot,
            group size=3 by 1,
            xlabels at=edge bottom,
            ylabels at=edge left,
            horizontal sep=0pt,
            vertical sep=0pt,
            /pgf/bar width=7pt,
        },
        ylabel={Top-1 Accuracy},
        ybar= \pgflinewidth,
        width=\linewidth,
        x tick label style={rotate=0, anchor=center},
        xticklabel style={yshift=-2mm,xshift={ifthenelse(\ticknum==2,0,0)}},
        xtick=data,
        xticklabels={{\color{ImageDark} \imnetShort},{\color{ImageDark} \inatShort},{\color{ImageDark} \placesThreeShort},{\color{VideoDark} \sthsthShort},{\color{VideoDark} \epicShort}},
        ymajorgrids=true,
        grid style=dotted,
        nodes near coords,
        every node near coord/.append style={
            font=\tiny,
            xshift=0cm,
            /utils/exec={\setbox0\hbox{\pgfmathprintnumber\pgfplotspointmeta}
            \pgfmathfloattomacro{\pgfplotspointmeta}{\F}{\M}{\E}
            \pgfmathsetmacro{\myanchor}{ifthenelse(\M*pow(10,\E-3)*2-\the\wd0>0,"east","west")}},
            rotate=90,anchor=\myanchor,/pgf/number format/.cd,fixed zerofill,precision=1
        },
        scale only axis,
        axis x line*=bottom,
        axis y line*=left,
        enlarge x limits = {abs=1},
        cycle list={
            draw=DarkGray,thick,fill=Gray!60,fill opacity=0.6\\
            draw=DarkGray,thick,fill=DarkGray!65,fill opacity=0.6\\
            draw=DarkGray,thick,fill=amber!60,fill opacity=1.0\\
        },
        legend columns=1,
        legend style={at={(0.095,1.2)}},
        height=0.15\textwidth,
        ymin=15,
    ]
        \nextgroupplot[
            xlabel=\vitB,
            width=\WdFirst\textwidth,
            ytick pos=left,
            xmin=1.5,
            ymin=10,
            ymax=90,
        ]
        \addplot  coordinates {
            (1, 83.4) %
            (2, 75.6) %
            (3, 58.6) %
            (4, 59.5) %
            (5, 21.5) %

        };
        \addplot  coordinates {
            (1, 81.1) %
            (2, 66.9) %
            (3, 57.4) %
            (4, 69.3) %
            (5, 39.9) %
        };
        \addplot  coordinates {
            (1, 82.8) %
            (2, 73.0) %
            (3, 58.2) %
            (4, 69.0) %
            (5, 39.6) %
        };

        \nextgroupplot[
            xlabel=\vitL,
            yticklabels={},  %
            width=\WdSecond\textwidth,
            ymajorticks=false,
            y axis line style={draw=none},
            ymin=10,
            ymax=90,
            xmax=4.5,
        ]
        \addplot coordinates {
            (1, 85.5) %
            (2, 80.3) %
            (3, 59.4) %
            (4, 57.7) %
            (5, 18.3) %
        };
        \addplot coordinates {
            (1, 81.7)  %
            (2, 70.5) %
            (3, 58.1) %
            (4, 73.2) %
            (5, 45.1) %
        };
        \addplot coordinates {
            (1, 84.7) %
            (2, 78.1) %
            (3, 59.4) %
            (4, 73.4) %
            (5, 44.3) %
        };

        \legend{\mae, \videomae, {\bf \OURS}}
    \end{groupplot}
\end{tikzpicture}%
    }
    \caption{{\bf \OURS on {\color{ImageDark}image} and {\color{VideoDark}video} downstream tasks.}
      We finetune the \mae, \videomae, and \OURS models on image and video benchmarks.
      We use the \vit architecture with two model sizes: \vitB and \vitL.
      \mae has poor video recognition performance while \videomae's performance drops on image datasets.
      \OURS pretraining generalizes to both benchmarks.
      All models are trained for 800 epochs on the pretraining datasets.
      The image-only MAE model is inflated~\cite{carreira2017quo} to apply MAE to video recognition tasks.
      The input image is replicated to apply \videomae to image recognition benchmarks.
      }
    \label{fig:compare_all_maes}
\end{figure*}
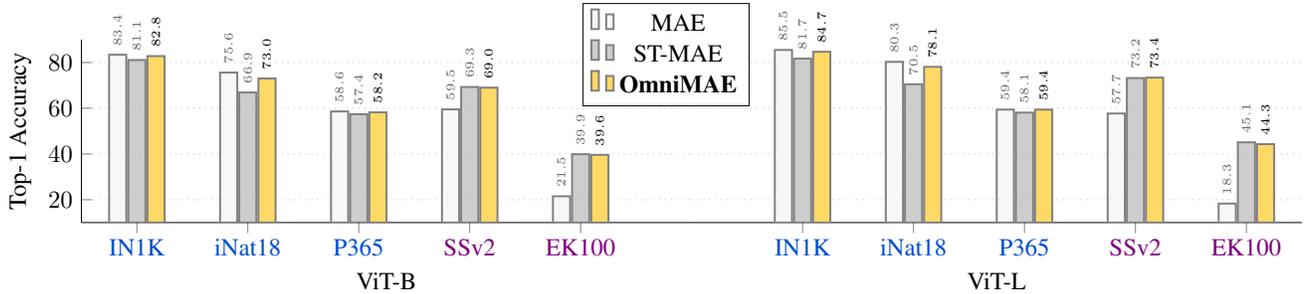

\textbf{Pretraining.}
We convert the input into $\totalPatch$ spatio-temporal patches and randomly mask $\totalMaskPatch$
of them.
Following~\cite{he2021masked}, we feed the non-masked $\totalPatch\!-\!\totalMaskPatch$ patches (and
their positions) to the encoder to produce per-patch embeddings. The encoder output is concatenated
with $\totalMaskPatch$ copies of a learnable `mask token' to obtain
$\totalPatch$ embeddings. We add the appropriate positional encoding to
the $\totalPatch$ embeddings and feed them into the decoder, which outputs $\totalPatch\!\times\! t
\times\! h\!\times\! w\!\times\! 3$ pixel values.

\textbf{Loss function and optimization.}
We minimize the reconstruction error between the decoder predictions and the input pixel values. The
input pixel values are normalized to zero mean and unit variance~\cite{ioffe2015batch} to get the
targets for the loss.
We minimize the $\ell_{2}$ distance between the predictions and targets over the $\totalMaskPatch$
masked patches. At each point in the training, we sample a mini-batch of either images or video and
compute the loss using the decoder predictions. We study the effect of different mini-batch
construction strategies on the overall performance and speed of training in~\cref{sec:ablations}.

\par \noindent \textbf{Dataset sample replication.} Since we operate on a small subset of the
input patches ($M\!\ll\!N$), we find that our data reading speed is unable to keep up with the
number of data points we can theoretically process in a single optimization step.
This issue is even more pronounced for videos, where we spend significant compute to read and decode a full video,
only to discard $>\!90\%$ of it.
Inspired from the success of RepeatAugment, which shows that
replicating samples within a mini-batch is an effective technique to improve
generalization~\cite{hoffer2020augment,berman2019multigrain,caron2020unsupervised},
we replicate a single data point and apply different masks to it each time.
Even with replicated samples, the non-masked input to the model is different due to random cropping and masking being different across the replicas.
We show in~\cref{sec:ablations} that sample
replication leads to gains in runtime without affecting the final transfer accuracy.

\subsection{Implementation Details}
We note some of the salient implementation details and provide the complete details in~\cref{appendix:impl_details}.

{\noindent \bf Architecture.} We use the ViT~\cite{dosovitskiy2020image}
architecture for the omnivorous encoder and experiment with its \vitB, \vitL, and \vitH variants. We do not
use the \texttt{[CLS]} token in the ViT models, yielding a small improvement in runtime without loss in performance.
We use a Transformer decoder with 4 layers (8 for \vitH) of 384, 512, and 512 embedding dimensions for \vitB, \vitL,
and \vitH, respectively. The decoder outputs the RGB colors for the pixels in all the input patches.
We use sinusoidal positional encoding~\cite{vaswani2017attention} for the patches in both the encoder and
the decoder.

\par \noindent \textbf{Training details.} We train our models with a mini-batch size of $2048$.
We resize the input images and videos spatially to $224\!\times\!224$
pixels. For videos, we sample a clip of $T\!=\!16$ frames at 6 FPS.  %
We use a patch size of $2\!\times\!16\!\times\!16$ for \vitB and \vitL, and $2\!\times\!14\!\times\!14$ for \vitH. Images are replicated temporally to meet the patch size.

\par \noindent \textbf{Masking the input.} Compared to prior work, we use extremely high
masking for pretraining and only 10\% and 5\% of the image and video patches are fed to the
encoder for pretraining. We uniform randomly mask the patches for images and videos and ablate the masking hyperparameters
in~\cref{sec:ablations}.

\section{Experiments}
\label{sec:experiments}

\begin{figure*}
    \centering
    \includegraphics[width=0.9\linewidth]{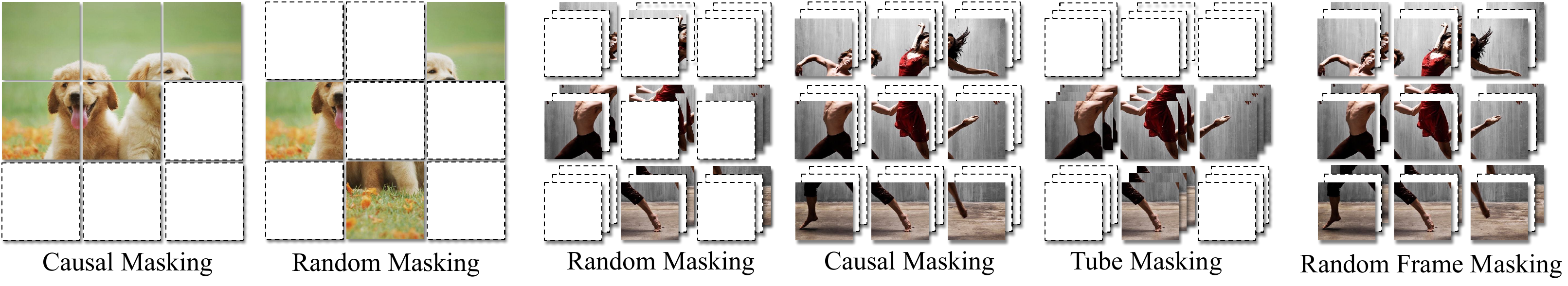}
    \caption{{\bf Different types of masking} for images (left two) and videos. Causal and tube
    masking use the data's spatio-temporal structure. Random frame masking randomly
    masks frames in a video. Random masking randomly masks patches and is used by default for
    \OURS.}
    \label{fig:masking_types}
\end{figure*}

{\noindent \bf Pretraining data.} We pretrain representations by jointly training on images from the \imnet
(\imnetShort)~\cite{ILSVRC15} dataset and videos from the \sthsth
(\sthsthShort)~\cite{goyal2017something} dataset. We choose the \sthsthShort dataset over
web video datasets like \kinetics (\kineticsShort)~\cite{kay2017kinetics} for reproducibility, and given \sthsthShort's challenging nature requiring temporal reasoning.

{\noindent \bf Transfer learning datasets.} We consider two image datasets: \inat
(\inatShort)~\cite{iNaturalist}, a popular fine-grained recognition dataset, and \placesThree
(\placesThreeShort)~\cite{Places205}, a scene recognition benchmark. For video datasets, we focus on the popular
\kineticsShort action recognition benchmark, and \epic
(\epicShort)~\cite{damen2021rescaling}, an egocentric action recognition dataset. We report
the top-1 classification accuracy for all transfer datasets. The details about the datasets
can be found in~\cref{appendix:datasets}.

\subsection{\OURS on Image and Video Recognition}
\label{sec:joint_image_video_experiments}
We now evaluate the capabilities of \OURS's representations on both image and video recognition tasks.

{\noindent \bf Baselines.}
We compare \OURS's representations trained jointly on images and videos to \mae, trained solely on images~\cite{he2021masked}. Additionally, we develop a video-only baseline, \vmaeFull (\videomae), by training \OURS only on videos.
For a fair comparison, all models are pretrained for 800 epochs on their respective datasets.
The image-only \mae model is inflated~\cite{carreira2017quo} for finetuning on video tasks, see~\cref{appendix:transfer_details} %
for details.

{\noindent \bf Observations.}
\cref{fig:compare_all_maes} presents the evaluation results for all the models on the image and video benchmarks.
The modality specific MAE and \videomae models achieve good transfer performance when transferring to datasets which match the pretraining modality.
However, both models show a degradation in performance when transferring to a different visual modality.
MAE has lower video recognition performance, especially on egocentric videos from \epicShort where the smaller \vitB \mae model is 18.4\% worse compared to \videomae.
Similarly, compared to \mae, \videomae is 8.7\% worse on the fine-grained classification task of \inatShort.
With a large model, MAE's performance on the image recognition benchmarks improves but the cross-modality performance degrades further.
On \epic the large MAE model is more than 25\% worse than \videomae.

When transferred to both image and video recognition tasks, \OURS performs favorably to the single modality baselines that use exactly the same architecture.
\OURS also uses the same finetuning recipe as the single modality baselines.
\OURS matches the video classification performance of \videomae and the image classification performance of MAE
for both \vitB and \vitL, with \OURS's performance improving on both image and video recognition benchmarks with the larger model.
These results suggest that our unified model of images and videos serves as a
better pretraining across a wider variety of recognition tasks than single modality models.
\begin{table*}[t]
    \setlength{\tabcolsep}{2pt}
    \centering
    \begin{subtable}[b]{.18\linewidth}
        \centering
        \vspace{0pt}  %
        \resizebox{\linewidth}{!}{
            \begin{tabular}{cc | cc}
    \multicolumn{2}{c}{\bf Ratios} & \multicolumn{2}{c}{\bf Accuracy} \\
    {\color{ImageDark} \bf \imnetShort} & {\color{VideoDark} \bf \sthsthShort} & {\color{ImageDark} \bf \imnetShort} & {\color{VideoDark} \bf \sthsthShort} \\
    \midrule
    75\% & 75\% & 82.6 & 67.2 \\ %
    \colorrow 75\% & 90\% & 82.8 & 67.9 \\  %
    75\% & 95\% & 82.8 & 68.1 \\ %
    90\% & 75\% & 82.8 & 67.7 \\  %
    90\% & 90\% & 82.6 & 68.5 \\  %
    90\% & 95\% & {\bf 82.8} & {\bf 68.6} \\  %
\end{tabular}%

        }
        \caption{Masking ratios.}
        \label{tab:ablations_masking_ratio}
    \end{subtable}\hfill
    \begin{subtable}[b]{.2\linewidth}
        \centering
        \vspace{0pt}  %
        \resizebox{\linewidth}{!}{
            \begin{tabular}{cc | cc}
    \multicolumn{2}{c}{\bf Type} & \multicolumn{2}{c}{\bf Accuracy} \\
    {\color{ImageDark} \bf \imnetShort} & {\color{VideoDark} \bf \sthsthShort} & {\color{ImageDark} \bf \imnetShort} & {\color{VideoDark} \bf \sthsthShort} \\
    \midrule
    \colorrow Random & Tube &
    82.8  %
    &
    {\bf 67.9}  %
    \\
    Causal & Tube &
    82.1  %
    &
    67.3  %
    \\
    \midrule
    Random & Frame &
    {\bf 82.9}  %
    &
    65.8  %
    \\
    Random & Causal &
    82.8 %
    &
    64.4  %
    \\
    Random & Random &
    82.7  %
    &
    67.8  %
    \\
\end{tabular}%

        }
        \caption{Masking type.}
        \label{tab:ablations_masking_type}
    \end{subtable}\hfill
    \begin{subtable}[b]{.2\linewidth}
        \centering
        \vspace{0pt}  %
        \resizebox{\linewidth}{!}{
            \begin{tabular}{lr|cc}
    &\bf \texttt{\#}params & {\color{ImageDark} \bf \imnetShort} & {\color{VideoDark} \bf \sthsthShort} \\
    \midrule
    \colorrow Common & 7.9M & {\bf 82.8} %
    & {\bf 67.9}  %
    \\
    Separate & 33.0M & 82.6 %
    & 67.7 %
    \\
    \midrule
    $L$=2 & 3.8M & 82.7 %
    & 67.9
    \\
    \colorrow $L$=4 & 7.9M & {\bf 82.8} %
    & 67.9  %
    \\
    $L$=8 & 14.5M & {\bf 82.8} %
    & {\bf 68.0}
    \\
    \midrule
    \colorrow $d$=384 & 7.9M & {\bf 82.8} %
    & 67.9  %
    \\
    $d$=512 & 13.0M & 82.7 %
    & {\bf 68.1}
    \\
\end{tabular}%

        }
        \caption{Decoder capacity.}
        \label{tab:ablations_decoder}
    \end{subtable}\hfill
    \begin{subfigure}[b]{.35\linewidth}
        \centering
        \vspace{0pt}  %
        \resizebox{\linewidth}{!}{
            \begin{tikzpicture}
    \begin{axis}[
        xmax=3.5,
        xmode=log,
        log ticks with fixed point,
        xtick={0.25, 0.5, 1, 2, 3},
        xticklabels={3:1, 2:1, 1:1, 1:2, 1:3},
        height=1.8in,
        axis x line*=bottom,
        axis y line*=left,
        width=\linewidth,
        ylabel style = {align=center},
        ylabel={\imnetShort~\ref{pgf:data_ratio:img}},
        yticklabel style = {font=\small},
        xticklabel style = {font=\tiny},
        legend style={cells={align=left}, font=\tiny},
    ]
    \addplot[mark=\ImageMark, thick, ImageDark] plot coordinates {
        (0.25, 83.0) %
        (0.5, 83.0) %
        (1, 82.8)
        (2, 82.8) %
        (3, 82.7)%
    };\label{pgf:data_ratio:img}
    \end{axis}

    \begin{axis}[
        xmax=3.5,
        xmode=log,
        log ticks with fixed point,
        height=1.8in,
        xtick={0.25, 0.5, 1, 2, 3},
        axis x line=none,
        axis y line*=right,
        width=\linewidth,
        ylabel style = {align=center},
        ylabel={\sthsthShort~\ref{pgf:data_ratio:video}},
        yticklabel style = {font=\small},
        legend style={cells={align=left}, font=\tiny},
    ]
    \addplot[mark=\VideoMark, thick, VideoDark] plot coordinates {
        (0.25, 67.5)  %
        (0.5, 67.7) %
        (1, 68.3)
        (2, 68.7)  %
        (3, 69.1)  %
    };\label{pgf:data_ratio:video}
    \end{axis}
\end{tikzpicture}
        }
        \caption{Dataset ratio}
        \label{fig:graphs_repeat_sampling_dataset}
    \end{subfigure}
    \caption{{\bf Ablations.}
    The default setting for all ablations is highlighted in \colorrowtext{}.
    \textbf{(a)}
    We vary the masking ratios of the input for pretraining the model.
    This allows us to use an extremely high masking ratio of 90\% on images and 95\% on videos.
    \textbf{(b)} \OURS works well across different masking types.
    Random masking makes no assumptions about the input patches.
    Tube masking masks random patches at the same spatial location across all frames.
    Causal masking for images masks the `future' patches as determined by a raster left-to-right order.
    In videos, causal masking masks all the patches from future frames.
    Frame masking masks all patches for randomly selected frames.
    \textbf{(c)} Different decoder designs (common or separate) and capacities (number of layers $L$ and dimension $d$ of the MLP).
    A common decoder for both images and videos performs better than a separate decoder.
    Our final performance is robust to decoder capacity.
    \textbf{(d)} We vary the image/video dataset ratios by replicating the entire datasets by a factor.
    The number of training updates changes with such replication, and we observe that the model benefits from higher replication of videos.
    }
    \label{tab:ablations}
\end{table*}

{\noindent \bf Qualitative results.}
Following~\cite{he2021masked}, we re-train \OURS without normalizing the pixel targets to obtain easy to visualize RGB reconstructions.
We visualize the predictions in~\cref{fig:visualizations} using samples that are unseen during training: either \texttt{val} sets or different datasets altogether.
\OURS makes reasonable predictions on the in-distribution \imnet and \sthsthShort val sets, as well as the unseen, out-of-distribution \kineticsShort and \epicShort datasets.
As expected, the details in the reconstruction decrease when increasing the masking ratio.
However, even with 95\% masking, the model can reconstruct coarse details in the input, \eg, in \imnet, the model reconstructs the coarse structure of the flower, vehicle, dog \etc.
\begin{table*}[t]
     \centering
     \setlength{\tabcolsep}{2pt}
     \newcommand{\raiseboxht}[0]{-0.45}
     \newcommand{\raiseboxlastht}[0]{\raiseboxht}
     \newcommand{\vizcropht}[0]{8.1cm}
     \newcommand{\vizcroplastht}[0]{16.3cm}
     \newcommand{\percolwidth}[0]{0.48}
     \newcommand{\perimgcolwidth}[0]{0.06}  %
     \hspace{-0.15in}  %
     \resizebox{\linewidth}{!}{
     \begin{tabular}{ccc}
          & \color{VideoDark} \bf\sthsthShort & \color{VideoDark} \bf\kineticsShort \\
          \rotatebox[origin=c]{90}{\footnotesize Ref}
          & \raisebox{\raiseboxlastht\height}{\includegraphics[width=\percolwidth\linewidth, trim={0 0 0 \vizcroplastht{}},clip]{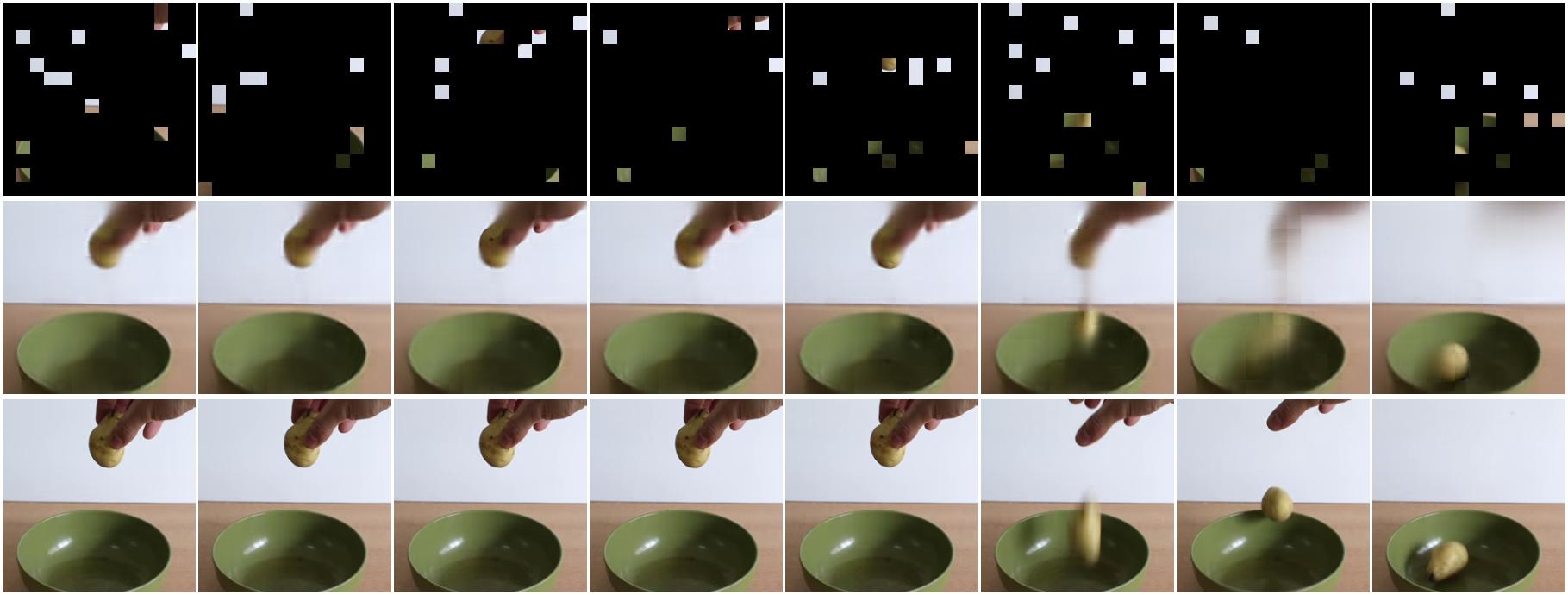}}
          & \raisebox{\raiseboxlastht\height}{\includegraphics[width=\percolwidth\linewidth, trim={0 0 0 \vizcroplastht{}},clip]{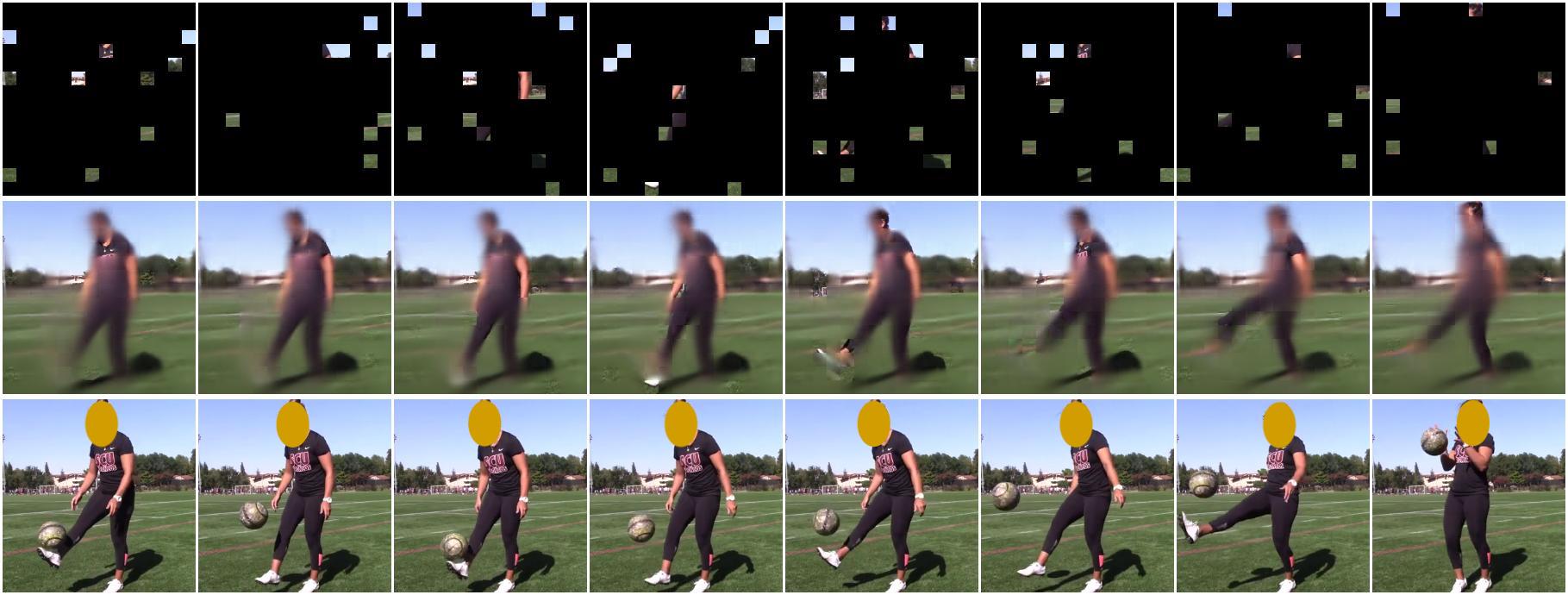}} \\
          \rotatebox[origin=c]{90}{\footnotesize 75\% mask}
          & \raisebox{\raiseboxht\height}{\includegraphics[width=\percolwidth\linewidth, trim={0 \vizcropht{} 0 0},clip]{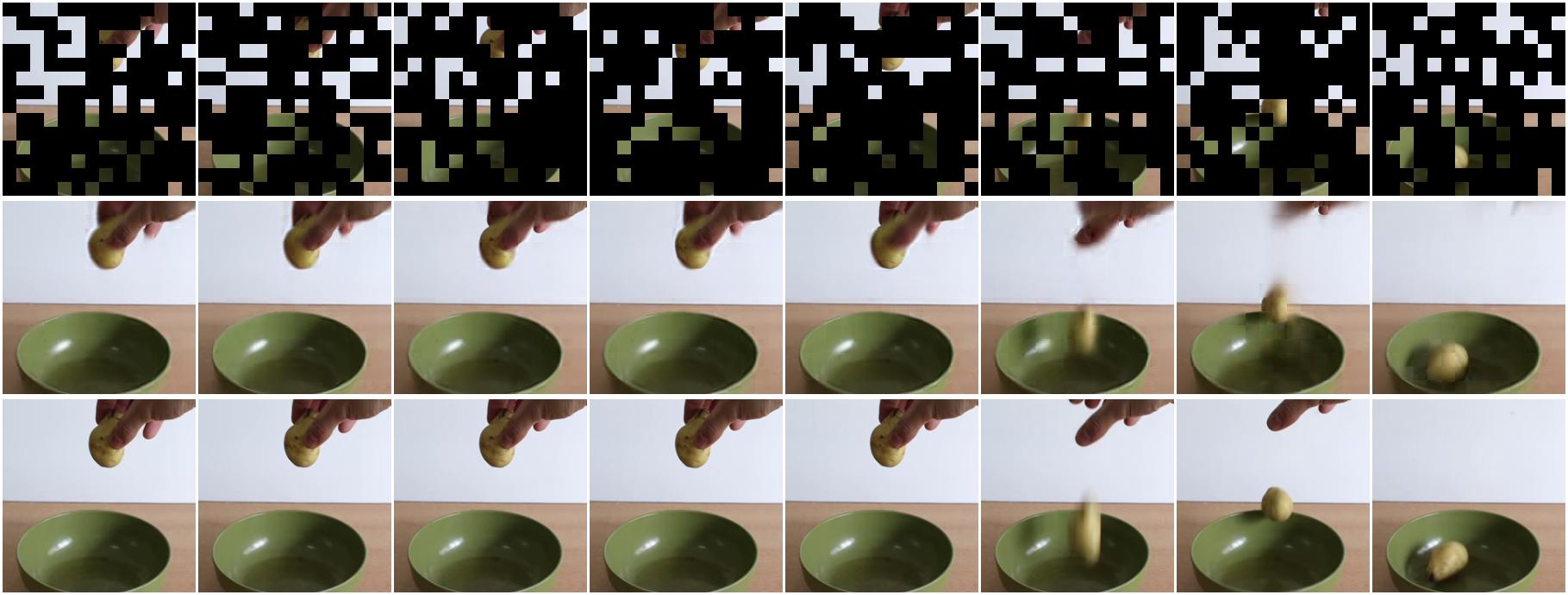}}
          & \raisebox{\raiseboxht\height}{\includegraphics[width=\percolwidth\linewidth, trim={0 \vizcropht{} 0 0},clip]{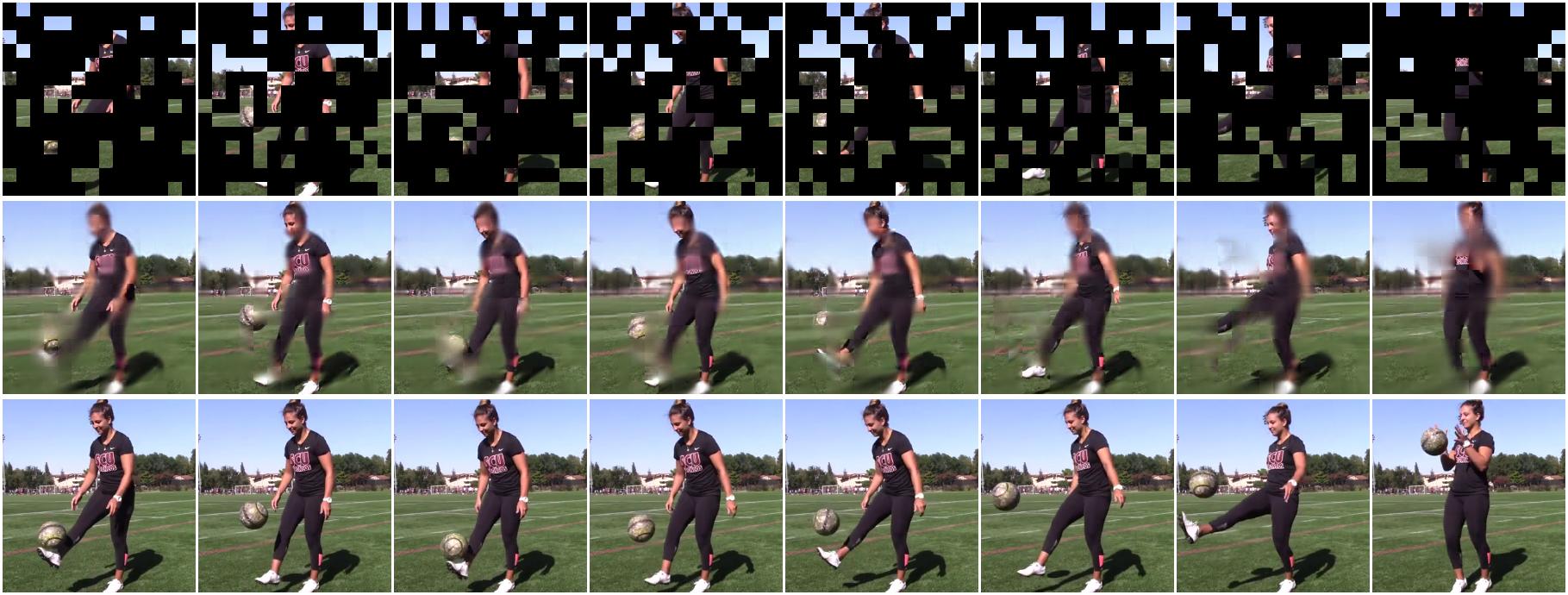}}
          \\
          \rotatebox[origin=c]{90}{\footnotesize 90\% mask}
          & \raisebox{\raiseboxht\height}{\includegraphics[width=\percolwidth\linewidth, trim={0 \vizcropht{} 0 0},clip]{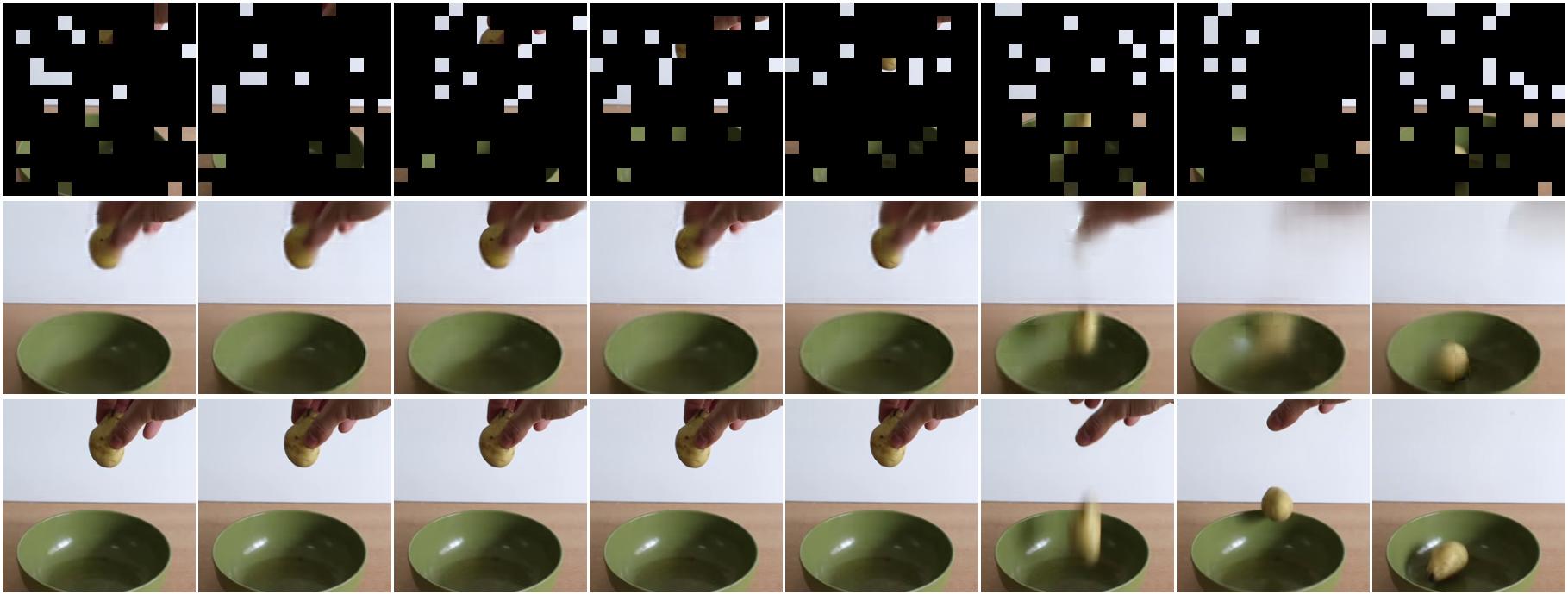}}
          & \raisebox{\raiseboxht\height}{\includegraphics[width=\percolwidth\linewidth, trim={0 \vizcropht{} 0 0},clip]{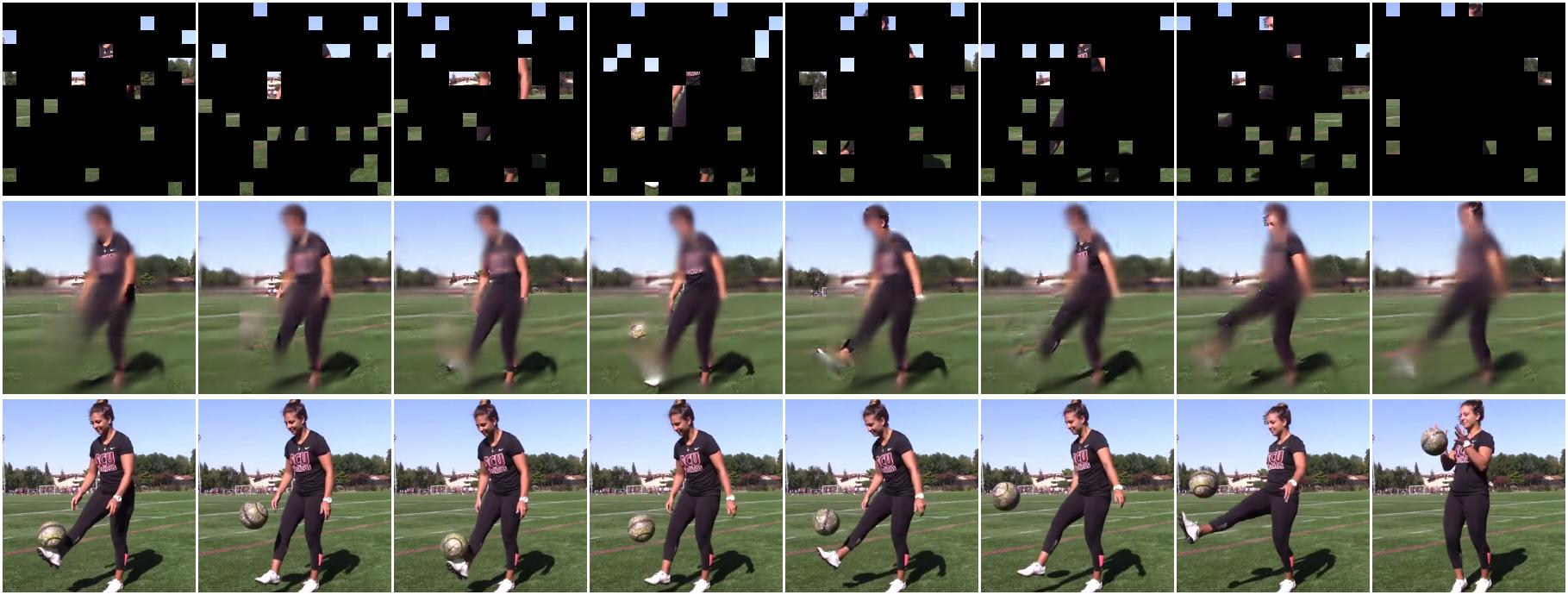}}
          \\
          \rotatebox[origin=c]{90}{\footnotesize 95\% mask}
          & \raisebox{\raiseboxht\height}{\includegraphics[width=\percolwidth\linewidth, trim={0 \vizcropht{} 0 0},clip]{figures/fig_vid_compare/mae_concat_ssv2_mr_95.jpg}}
          & \raisebox{\raiseboxht\height}{\includegraphics[width=\percolwidth\linewidth, trim={0 \vizcropht{} 0 0},clip]{figures/fig_vid_compare/mae_concat_k400_mr_95.jpg}}
          \\
          & \color{VideoDark} \bf\epicShort &  \color{ImageDark} \bf\imnetShort \\
          \rotatebox[origin=c]{90}{\footnotesize Ref}
          & \raisebox{\raiseboxlastht\height}{\includegraphics[width=\percolwidth\linewidth, trim={0 0 0 \vizcroplastht{}},clip]{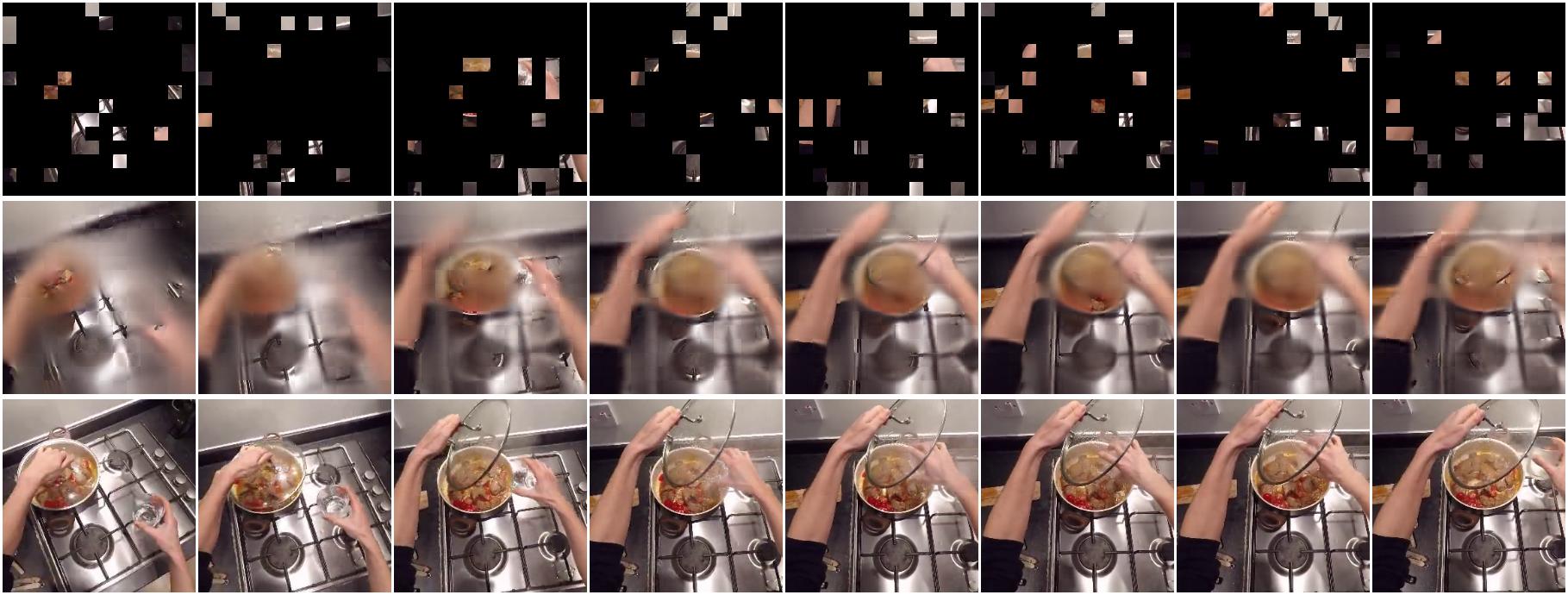}}
          & \raisebox{\raiseboxlastht\height}{
               \includegraphics[width=\perimgcolwidth\linewidth, trim={0 0 0 \vizcroplastht{}},clip]{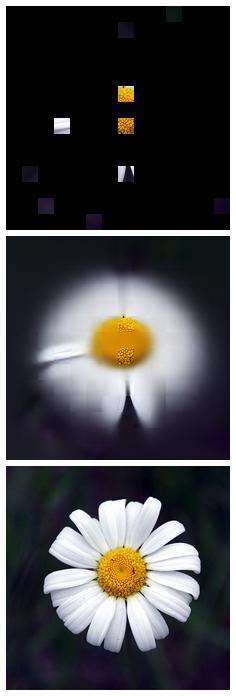} \hfill
               \includegraphics[width=\perimgcolwidth\linewidth, trim={0 0 0 \vizcroplastht{}},clip]{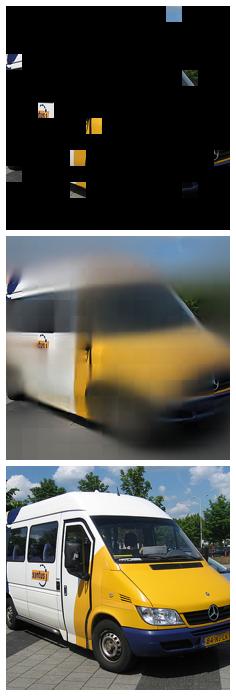} \hfill
               \includegraphics[width=\perimgcolwidth\linewidth, trim={0 0 0 \vizcroplastht{}},clip]{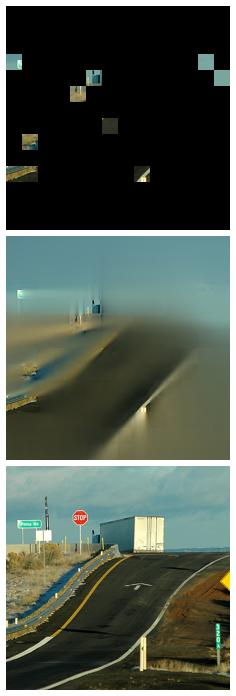} \hfill
               \includegraphics[width=\perimgcolwidth\linewidth, trim={0 0 0 \vizcroplastht{}},clip]{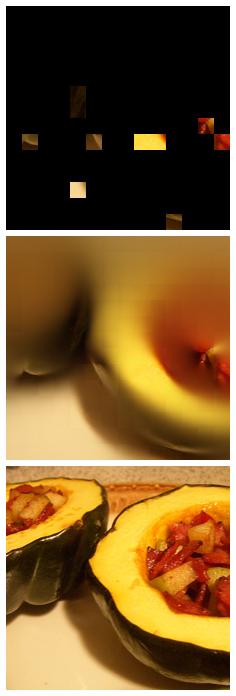} \hfill
               \includegraphics[width=\perimgcolwidth\linewidth, trim={0 0 0 \vizcroplastht{}},clip]{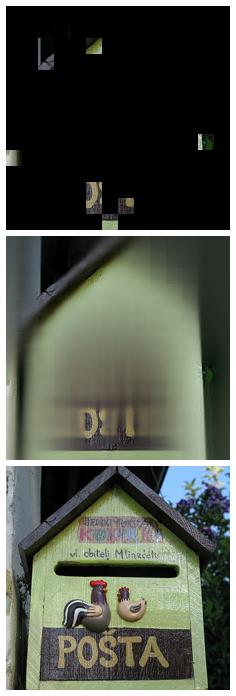} \hfill
               \includegraphics[width=\perimgcolwidth\linewidth, trim={0 0 0 \vizcroplastht{}},clip]{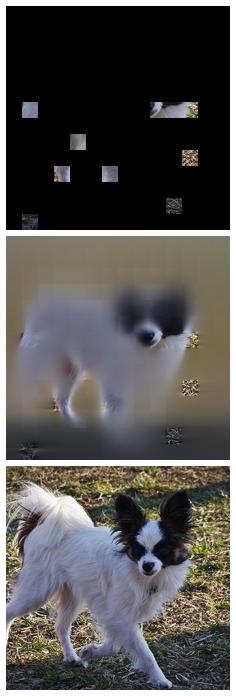} \hfill
               \includegraphics[width=\perimgcolwidth\linewidth, trim={0 0 0 \vizcroplastht{}},clip]{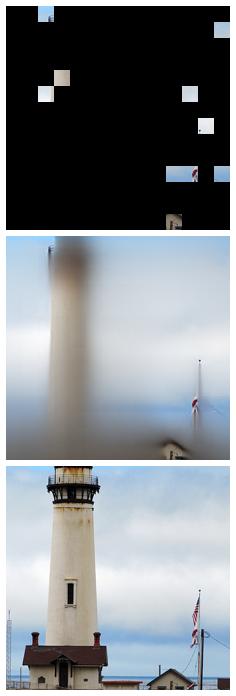}
          }
          \\
          \rotatebox[origin=c]{90}{\footnotesize 75\% mask}
          & \raisebox{\raiseboxht\height}{\includegraphics[width=\percolwidth\linewidth, trim={0 \vizcropht{} 0 0},clip]{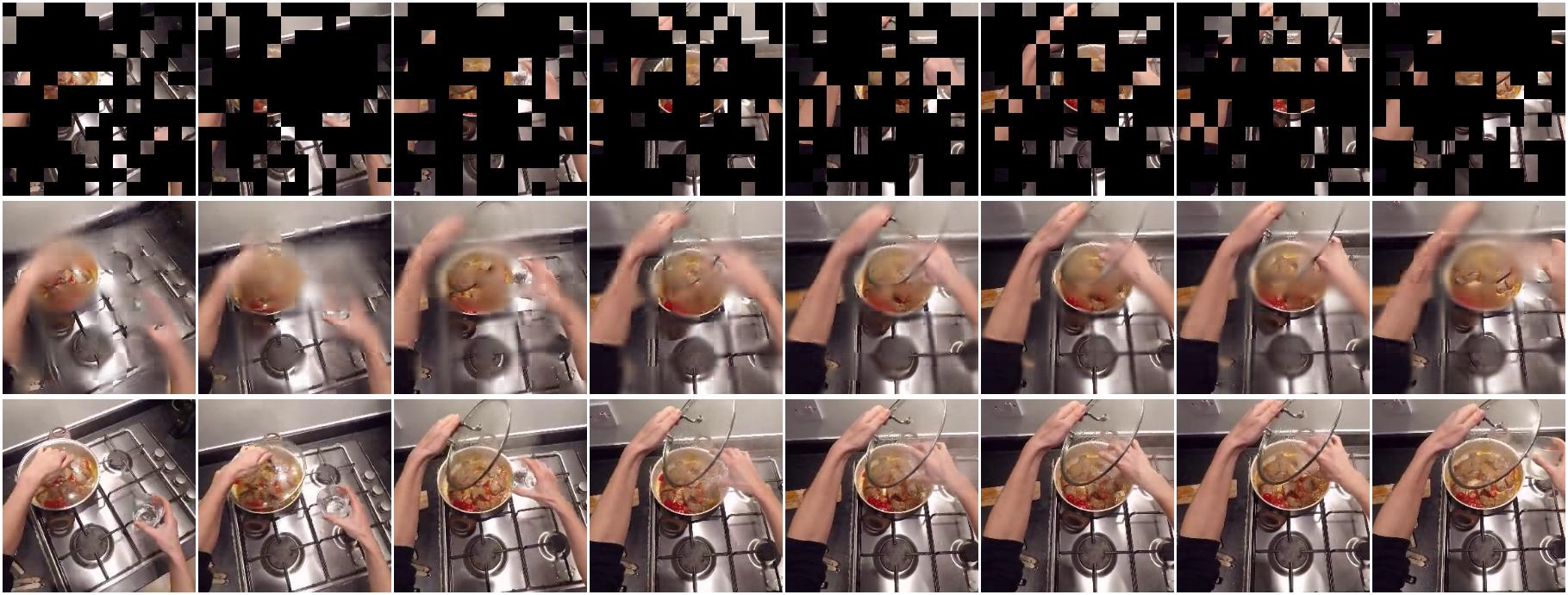}}
          & \raisebox{\raiseboxlastht\height}{
               \includegraphics[width=\perimgcolwidth\linewidth, trim={0 \vizcropht{} 0 0},clip]{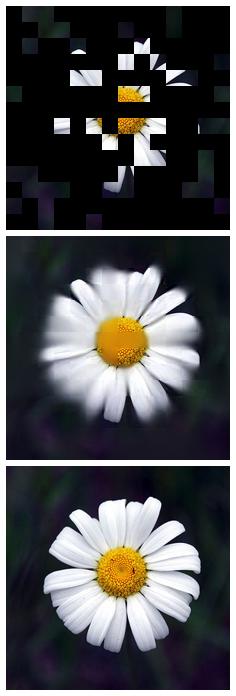} \hfill
               \includegraphics[width=\perimgcolwidth\linewidth, trim={0 \vizcropht{} 0 0},clip]{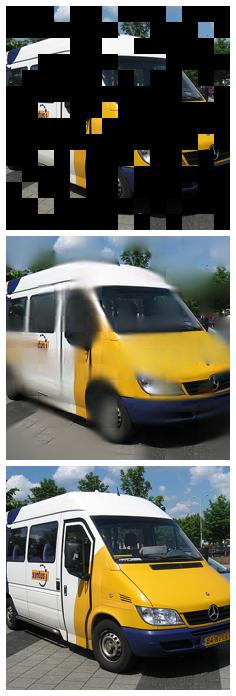} \hfill
               \includegraphics[width=\perimgcolwidth\linewidth, trim={0 \vizcropht{} 0 0},clip]{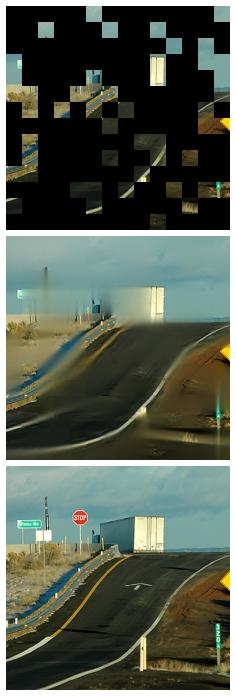} \hfill
               \includegraphics[width=\perimgcolwidth\linewidth, trim={0 \vizcropht{} 0 0},clip]{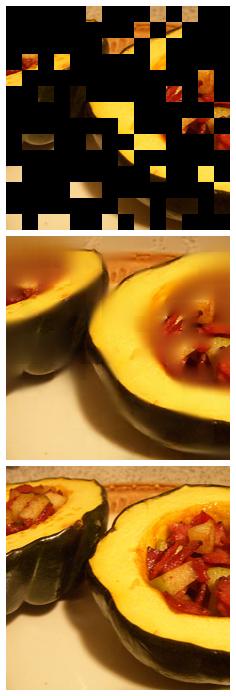} \hfill
               \includegraphics[width=\perimgcolwidth\linewidth, trim={0 \vizcropht{} 0 0},clip]{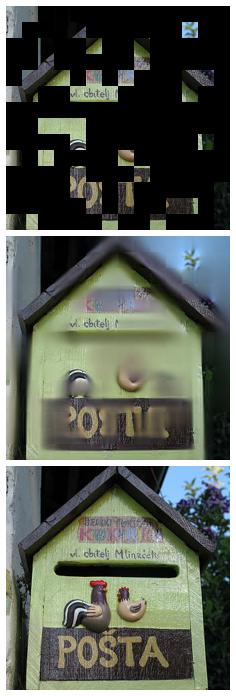} \hfill
               \includegraphics[width=\perimgcolwidth\linewidth, trim={0 \vizcropht{} 0 0},clip]{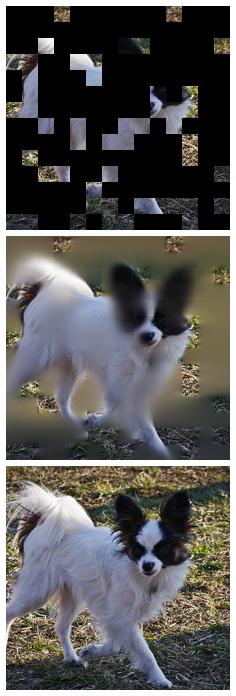} \hfill
               \includegraphics[width=\perimgcolwidth\linewidth, trim={0 \vizcropht{} 0 0},clip]{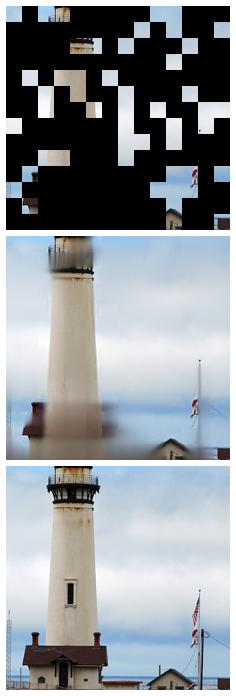}
          }
          \\
          \rotatebox[origin=c]{90}{\footnotesize 90\% mask}
          & \raisebox{\raiseboxht\height}{\includegraphics[width=\percolwidth\linewidth, trim={0 \vizcropht{} 0 0},clip]{figures/fig_vid_compare/mae_concat_epic_mr_90.jpg}}
          & \raisebox{\raiseboxlastht\height}{
               \includegraphics[width=\perimgcolwidth\linewidth, trim={0 \vizcropht{} 0 0},clip]{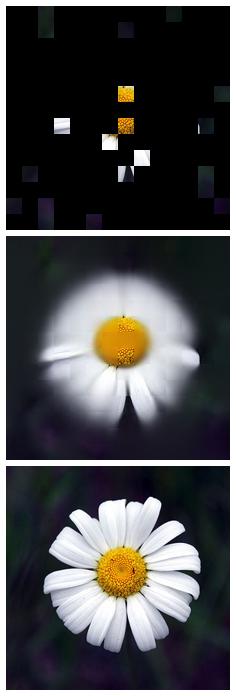} \hfill
               \includegraphics[width=\perimgcolwidth\linewidth, trim={0 \vizcropht{} 0 0},clip]{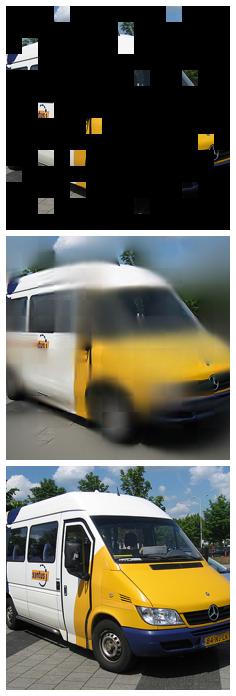} \hfill
               \includegraphics[width=\perimgcolwidth\linewidth, trim={0 \vizcropht{} 0 0},clip]{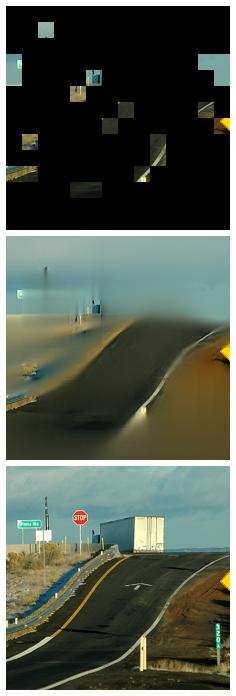} \hfill
               \includegraphics[width=\perimgcolwidth\linewidth, trim={0 \vizcropht{} 0 0},clip]{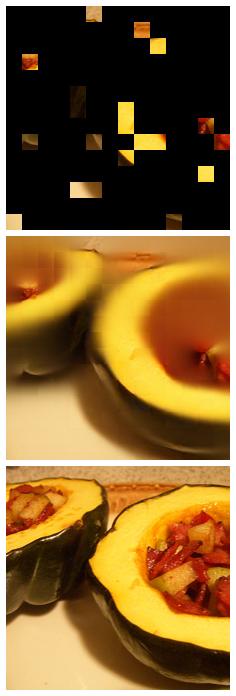} \hfill
               \includegraphics[width=\perimgcolwidth\linewidth, trim={0 \vizcropht{} 0 0},clip]{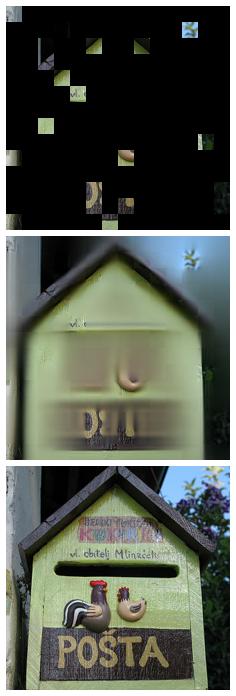} \hfill
               \includegraphics[width=\perimgcolwidth\linewidth, trim={0 \vizcropht{} 0 0},clip]{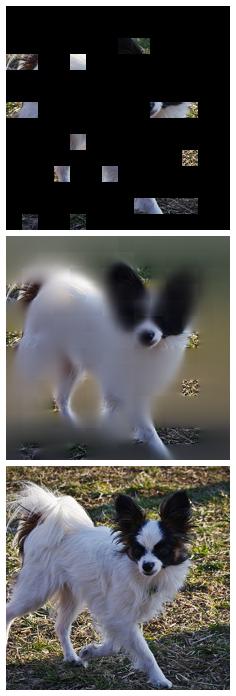} \hfill
               \includegraphics[width=\perimgcolwidth\linewidth, trim={0 \vizcropht{} 0 0},clip]{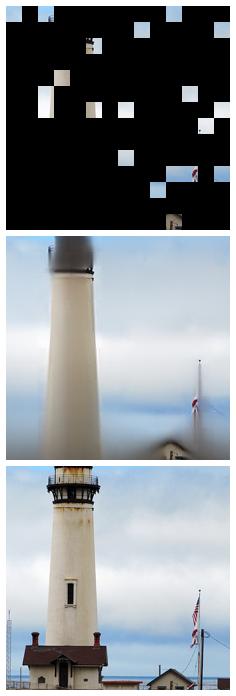}
          }
          \\
          \rotatebox[origin=c]{90}{\footnotesize 95\% mask}
          & \raisebox{\raiseboxht\height}{\includegraphics[width=\percolwidth\linewidth, trim={0 \vizcropht{} 0 0},clip]{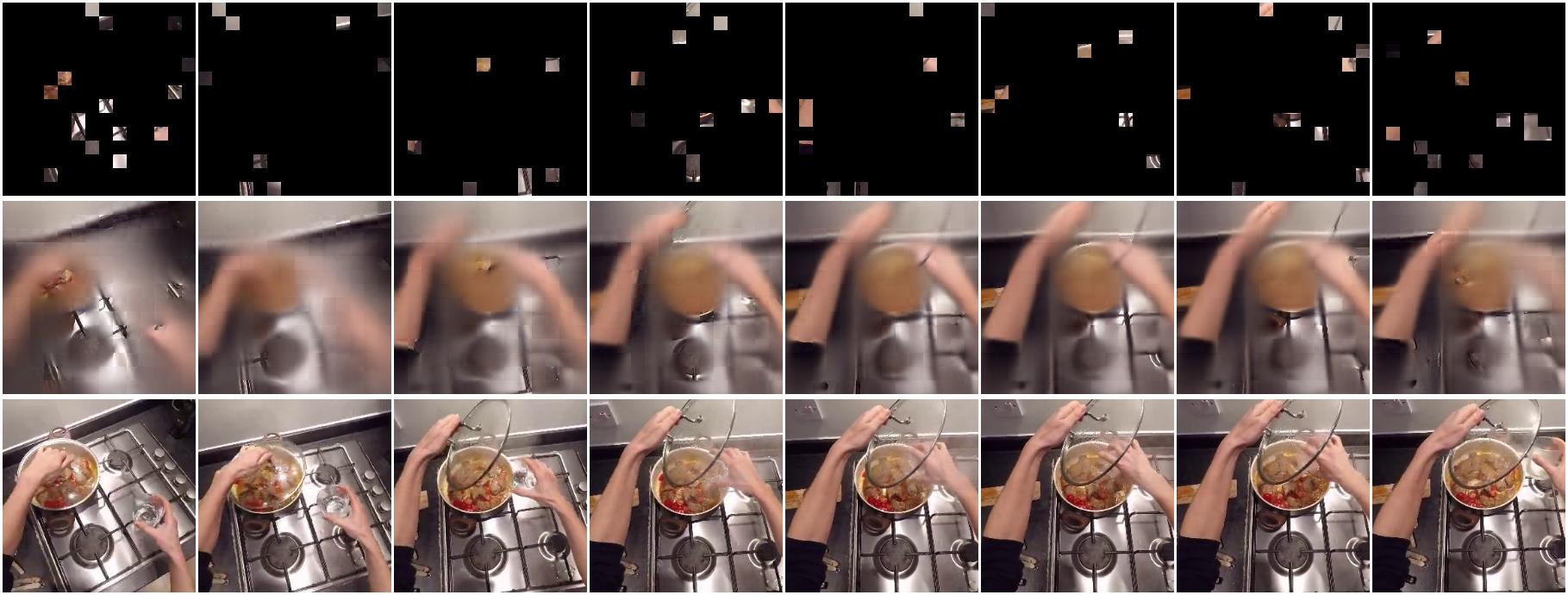}}
          & \raisebox{\raiseboxlastht\height}{
               \includegraphics[width=\perimgcolwidth\linewidth, trim={0 \vizcropht{} 0 0},clip]{figures/fig_img_compare/mae_concat_in1k_mr_95_1.jpg} \hfill
               \includegraphics[width=\perimgcolwidth\linewidth, trim={0 \vizcropht{} 0 0},clip]{figures/fig_img_compare/mae_concat_in1k_mr_95_2.jpg} \hfill
               \includegraphics[width=\perimgcolwidth\linewidth, trim={0 \vizcropht{} 0 0},clip]{figures/fig_img_compare/mae_concat_in1k_mr_95_3.jpg} \hfill
               \includegraphics[width=\perimgcolwidth\linewidth, trim={0 \vizcropht{} 0 0},clip]{figures/fig_img_compare/mae_concat_in1k_mr_95_4.jpg} \hfill
               \includegraphics[width=\perimgcolwidth\linewidth, trim={0 \vizcropht{} 0 0},clip]{figures/fig_img_compare/mae_concat_in1k_mr_95_5.jpg} \hfill
               \includegraphics[width=\perimgcolwidth\linewidth, trim={0 \vizcropht{} 0 0},clip]{figures/fig_img_compare/mae_concat_in1k_mr_95_6.jpg} \hfill
               \includegraphics[width=\perimgcolwidth\linewidth, trim={0 \vizcropht{} 0 0},clip]{figures/fig_img_compare/mae_concat_in1k_mr_95_7.jpg}
          }
          \\
     \end{tabular}}
     \vspace{0.07in}
     \captionof{figure}{{\bf Reconstruction visualizations} using \OURS on different {\color{VideoDark}video} and {\color{ImageDark}image} datasets.
     We show the model predictions for varying masking ratios of the input from 75\% to 95\% and the ground truth reference (Ref).
     \OURS is trained on \imnet and \sthsthShort but the predictions generalize to other datasets like \kineticsShort and \epicShort.
     Please see the supplement for video visualizations.
     }\label{fig:visualizations}
\end{table*}

\begin{table*}[t]
    \setlength{\tabcolsep}{2.5pt}
    \centering
    \resizebox{0.65\linewidth}{!}{%
\begin{tabu}{l l l cH cH cH cH cH cH}
    \bf Method & \bf Arch. & \bf Pretrain Data & \multicolumn{1}{c}{ \color{ImageDark} \bf\imnetShort} & & \multicolumn{1}{c}{ \color{ImageDark} \bf\inatShort} && \multicolumn{1}{c}{ \color{ImageDark} \bf\placesThreeShort} && \multicolumn{1}{c}{ \color{VideoDark} \bf\kineticsShort} && \multicolumn{1}{c}{ \color{VideoDark} \bf\sthsthShort} && \multicolumn{1}{c}{ \color{VideoDark} \bf\epicShort} \\
    \midrule
    \dino & \vitB & \imnetShort
    & 82.8 & ??
    & 72.6 & ?? %
    & -- & ??
    & \xmark & \xmark
    & \xmark & \xmark
    & \xmark & \xmark
    \\ %
    \ibot~\cite{zhou2021ibot} & \vitL & \imnetShort
    & 84.8 & ??
    & -- & ??
    & -- & ??
    & \xmark & \xmark
    & \xmark & \xmark
    & \xmark & \xmark
    \\ %
    \mae~\cite{he2021masked} & \vitB & \imnetShort
    & 83.6 & ??
    & 75.4 & ?? %
    & 57.9 & ?? %
    & \xmark & \xmark
    & \xmark & \xmark
    & \xmark & \xmark
    \\
    \mae~\cite{he2021masked} & \vitL & \imnetShort
    & 85.9 & ??
    & 80.1 & ?? %
    & 59.4 & ?? %
    & \xmark & \xmark
    & \xmark & \xmark
    & \xmark & \xmark
    \\
    \mae~\cite{he2021masked} & \vitH & \imnetShort
    & 86.9 & ??
    & 83.0 & ?? %
    & 59.8 & ?? %
    & \xmark & \xmark
    & \xmark & \xmark
    & \xmark & \xmark
    \\
    BEiT~\cite{bao2021beit} & \vitB & \imnetShort
    & 83.4 & ??
    & 72.3 & ??
    & -- & ??
    & \xmark & \xmark
    & \xmark & \xmark
    & \xmark & \xmark
    \\ %
    BEiT~\cite{bao2021beit} & \vitL & \imnetShort
    & 85.2 & ??
    & -- & ??
    & -- & ??
    & \xmark & \xmark
    & \xmark & \xmark
    & \xmark & \xmark
    \\ %
    \midrule
    VIMPAC~\cite{tan2021vimpac} & \vitL/2\textsuperscript{\textdagger} & HowTo100M%
    & \xmark & \xmark
    & \xmark & \xmark
    & \xmark & \xmark
    & 77.4 & --
    & 68.1 & --
    & -- & --
    \\ %
    \maskedFeatV~\cite{wei2021masked} & MViT-v2-L & \kineticsShort
    & \xmark & \xmark
    & \xmark & \xmark
    & \xmark & \xmark
    & 84.3 & 96.3
    & -- & --
    & -- & --
    \\ %
    BEVT~\cite{wang2021bevt} & \swinB & \imnetShort + \kineticsShort
    & \xmark & \xmark
    & \xmark & \xmark
    & \xmark & \xmark
    & 81.1 & --
    & 71.4 & --
    & -- & --
    \\ %
    \midrule
    \OURS & \vitB & \imnetShort + \kineticsShort
    & 82.9 & ?? %
    & 74.2 & ?? %
    & 58.5 & ?? %
    & 80.8 & ?? %
    & 69.0 & ?? %
    & 40.3 & ?? %
    \\
    \arrayrulecolor{DarkGray!60}
    \midrule
    \arrayrulecolor{Black}
    \OURS & \vitB & \imnetShort + \sthsthShort
    & 83.0 & ?? %
    & 74.0 & ?? %
    & 58.4 & ?? %
    & 80.6 & ?? %
    & 69.5 & ?? %
    & 39.3 & ?? %
    \\
    \OURS & \vitL & \imnetShort + \sthsthShort
    & 85.2 & ?? %
    & 79.6 & ?? %
    & 59.3 & ?? %
    & 84.0 & ?? %
    & 74.2 & ?? %
    & 45.1 & ?? %
    \\
    \OURS & \vitH & \imnetShort + \sthsthShort
    & 86.6 & ?? %
    & 83.2 & ?? %
    & 60.1 & ?? %
    & 85.4 & ?? %
    & 75.5 & ?? %
    & 48.0 & ?? %
    \\
    \midrule
    \multicolumn{9}{l}{\em Concurrent methods specialized for videos} \\
    \rowfont{\color{DarkGray!70}} VideoMAE-16f~\cite{tong2022videomae} & \vitL & \sthsthShort
    & \xmark & \xmark
    & \xmark & \xmark
    & \xmark & \xmark
    & -- & --
    & 74.2 & 94.7
    & -- & -- \\
    \rowfont{\color{DarkGray!70}} VideoMAE-16f~\cite{tong2022videomae} & \vitL & \kineticsShort
    & \xmark & \xmark
    & \xmark & \xmark
    & \xmark & \xmark
    & 84.7 & 96.5
    & -- & --
    & -- & -- \\
    \rowfont{\color{DarkGray!70}} MAE-Video~\cite{feichtenhofer2022masked} & \vitH & \kineticsShort
    & \xmark & \xmark
    & \xmark & \xmark
    & \xmark & \xmark
    & 85.1 & 96.6
    & 74.1 & 94.5
    & -- & -- \\
\end{tabu}%
}
\vspace{0.05in}
\caption{
    {\bf Comparing \OURS with prior self-supervised methods} on image and video recognition.
    Prior work trains a specialized model for a particular visual modality, sometimes using specialized architectures.
    Our single \OURS model is pretrained jointly on images and videos for 1600 epochs (\vitH for 2400) and performs competitively across all benchmarks while using a simple architecture and pretraining method, even outperforming concurrent work specialized for videos.
    \textsuperscript{\textdagger}\vitL with half the MLP embedding dimension
}
\label{tab:sota}

\end{table*}

\begin{figure}[!b]
    \centering
    \begin{subfigure}[b]{.5\linewidth}
        \centering
        \vspace{0pt}  %
        \resizebox{\linewidth}{!}{
            \begin{tikzpicture}
    \begin{axis}[
        xmode=log,
        log ticks with fixed point,
        xtick={1,2,4,8},
        height=1.5in,
        ymin=0.9, ymax=1,
        axis y line*=right,
        axis x line=none,
        width=\linewidth,
        ylabel style = {align=center},
        ylabel={Time/epoch \ref{pgf:repeat_sample:runtime_imnet}},
        yticklabel style = {font=\small}
    ]
    \addplot[ybar, ybar legend, black, fill=gray!10] %
    coordinates {
        (1, 1.0)
        (2, 0.9850746268656716)
        (4, 0.9680170575692963)
        (8, 0.9552238805970149)
    };\label{pgf:repeat_sample:runtime_imnet}
    \end{axis}

    \begin{axis} [
        ylabel={\imnetShort acc \ref{pgf:repeat_sample:perf_imnet}},
        ymin=80, ymax=85,
        axis x line*=bottom,
        axis y line*=left,
        legend pos=north east,
        xmode=log,
        log ticks with fixed point,
        xtick={1,2,4,8},
        height=1.5in,
        width=\linewidth,
        legend style={cells={align=left}, font=\small},
        label style={font=\small},
        tick label style={font=\small},
    ]
    \addplot[mark=\ImageMark,thick,ImageDark] plot coordinates {
        (1, 82.7)
        (2, 82.8)
        (4, 82.8)
        (8, 82.6)
    };
    \label{pgf:repeat_sample:perf_imnet}
    \end{axis}
\end{tikzpicture}
        }
        \caption{Image sample replication}
        \label{fig:graphs_repeat_sampling_image}
    \end{subfigure}\hfill
    \begin{subfigure}[b]{.5\linewidth}
        \centering
        \vspace{0pt}  %
        \resizebox{\linewidth}{!}{
            \begin{tikzpicture}
    \begin{axis}[
        xmode=log,
        log ticks with fixed point,
        xtick={1,2,4,8},
        height=1.5in,
        ymin=0.75, ymax=1,
        axis y line*=right,
        axis x line=none,
        width=\linewidth,
        ylabel style = {align=center},
        ylabel={Time/epoch \ref{pgf:repeat_sample:runtime}},
        yticklabel style = {font=\small},
        legend style={cells={align=left}, font=\tiny},
    ]
    \addplot[ybar, ybar legend, black, fill=gray!10] %
    coordinates {
        (1, 1.0)
        (2, 0.9253731343283582)
        (4, 0.8635394456289979)
        (8, 0.8272921108742004)
    };\label{pgf:repeat_sample:runtime}
    \end{axis}

    \begin{axis} [
        ylabel={\sthsthShort acc \ref{pgf:repeat_sample:perf}},
        ymin=65, ymax=70,
        axis x line*=bottom,
        axis y line*=left,
        legend pos=north east,
        xmode=log,
        log ticks with fixed point,
        xtick={1,2,4,8},
        height=1.5in,
        width=\linewidth,
        legend style={cells={align=left}, font=\tiny},
        label style={font=\small},
        tick label style={font=\small},
    ]
    \addplot[mark=\VideoMark,thick,VideoDark] plot coordinates {
        (1, 68.0) %
        (2, 68.1) %
        (4, 68.0)
        (8, 67.9)
    };
    \label{pgf:repeat_sample:perf}
    \end{axis}
\end{tikzpicture}
        }
        \caption{Video sample replication}
        \label{fig:graphs_repeat_sampling_video}
    \end{subfigure}\hfill
    \caption{
    \textbf{Sample replication.} We study the effect of repeating samples while training our model.
    In each case, we repeat a sample $n$ times within a mini-batch while fixing the overall mini-batch size and training updates.
    Replication leads to improved training speeds, especially on video without affecting the final performance.
    }
\end{figure}
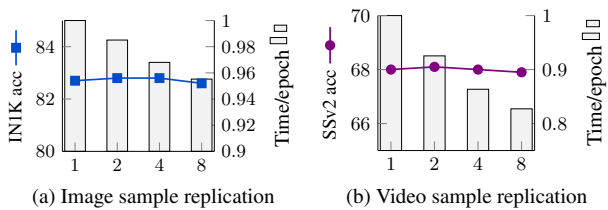

\subsection{Analyzing \OURS}
\label{sec:ablations}

We train the \vitB
architecture jointly on the \imnet and the \sthsthShort datasets for 800 epochs, where an epoch
involves training over all samples from both the datasets. For these analysis experiments, we use
the default masking hyperparameters of MAE for images (75\%).
For \vmae, due to redundancy across frames, we use ``tube'' dropping (\ie dropping all patches at a given spatial location over time) with high masking ratio (90\%)
All
hyperparameters for these experiments are in~\cref{appendix:ablation_details}. We evaluate all the
pretrained models on \imnet and \sthsthShort and report the top-1 classification accuracies
in~\cref{tab:ablations}. Please see~\cref{sec:ablations_appdx} for additional ablations.

{\noindent \bf Extreme masking.}
We vary the ratio of masked patches in the input while training the model.
We observe that videos benefit from significantly higher amounts of masking than images. Since
video frames have a lot of redundant information, the resulting spatio-temporal patches are also
redundant and thus higher masking leads to better self-supervised models. Unlike our experiments,
prior work found that extremely high masking ratios lead to degradation in performance, for instance
MAE~\cite{he2021masked} saw a significant degradation in performance when masking more than 75\%
of the patches. %
However, as we show in~\cref{tab:ablations_masking_ratio}, \OURS models can use extremely high masking of the input while
learning good representations.

{\noindent \bf Reduced compute due to extreme masking.} Our models trained with 90\% masking
on images and 95\% masking on videos yield good performance, while being trained with just 19 and 78
unmasked image and video patches respectively, for a patch size of $2\!\times\!16\!\times\!16$.
As we follow~\cite{he2021masked} and only pass
unmasked patches to the encoder and have a lightweight decoder, extreme masking leads to a dramatically
lower computational cost for training the encoder, and consequently the model as a whole.
Compared to using all the patches, our masked autoencoder uses $5.9\times$ and $7.8\times$ fewer
FLOPS for \vitB on images and videos respectively, $7.1\times$ and $11.6\times$ for \vitL, and
$7.2\times$ and $11.3\times$ for \vitH.
Compared to MAE, on \vitB, \vitL and \vitH, our higher masking leads to $1.8\times$, $2.0\times$ and
$2.0\times$ fewer FLOPS on images. On videos, compared to some concurrent works~\cite{tong2022videomae,feichtenhofer2022masked}
that use 90\% masking, we obtain $1.3\times$, $1.5\times$
and $1.4\times$ fewer FLOPS for \vitB/L/H. Given the compute savings and strong performance, we choose
90\% and 95\% masking for images and videos respectively for our final models.

{\noindent \bf Type of masking.} We study the effect of the type of masking used in training
our models. The different types of masking are illustrated in~\cref{fig:masking_types}. We
experiment with Random masking which masks patches in the image or video randomly. For videos, we
experiment with two types of masking that exploit the temporal structure. In Tube masking, we mask
random patches at the same spatial location across all the frames. In Causal masking, we use a raster order masking, akin to generative image models~\cite{chen2020generative,parmar2018image}, \ie the patches in the top-left of the image, and earlier frames of the video, are kept while the rest are masked.
Finally, in Frame masking, we randomly mask some frames in the video, while keeping all patches from the unmasked frames.
As seen in~\cref{tab:ablations_masking_type}, Random and Tube masking perform comparably well on both image and video tasks. We find that in case of Causal masking, the prediction task becomes exceedingly tough due to the uncertainity of the future, and in case of Frame masking, it becomes relatively easy due to the high redundancy of pixel values across frames. Hence in both these cases, the representation learned does not perform as well on video recognition tasks. Given the simplicity of random masking, we use that for both modalities.

{\noindent \bf Decoder architecture.} Since image and video prediction tasks may require
specialized parameters, we study two settings: (1) the decoder parameters are shared for image and
video pixel prediction; (2) two separate decoders are used for image and video prediction. For the latter
we use our default decoder setting for videos (4 layers/384-D), however a larger 8-layer/512-D decoder for images as proposed
in \mae~\cite{he2021masked}.
The results
in~\cref{tab:ablations_decoder} show that using a shared decoder for both image and video prediction
leads to better transfer learning accuracy.
A shared decoder also offers a $4\times$ reduction in the number of parameters compared to using
separate decoders while also being simpler to implement.

{\noindent \bf Decoder capacity.} In~\cref{tab:ablations_decoder}, we vary the decoder
capacity and measure its impact on the final transfer learning accuracy. To change the decoder
capacity, we vary the number of Transformer layers $L$ used and the dimension $d$ of the MLP used in
the Transformer layers. Overall, the final transfer learning accuracy is robust to decoder capacity.
A shallow decoder of 4 layers (8 for \vitH) offers a good trade-off between the decoder size and final accuracies.

{\noindent \bf Dataset ratios.} Since \imnet and \sthsthShort have a different number of
samples, we study the effect of varying the ratio of the datasets used in training. When increasing
the ratio of the datasets, we measure a single epoch as training over the new oversampled set of
samples from the datasets.
We vary the relative ratio for \imnet and \sthsth by replicating only one dataset
at a time as shown in~\cref{fig:graphs_repeat_sampling_dataset}. We observe that increasing the
relative dataset ratio has a positive effect on the final transfer accuracy for the oversampled
dataset. This is expected as oversampling a dataset proportionally increases its corresponding
parameter updates, thus making the representation better tuned for the dataset. We also observe that
oversampling \sthsth twice as much as \imnet leads to an improvement for the video transfer accuracy
with no drop in the image transfer performance. Hence, \OURS is robust to changes
in dataset sizes of individual modalities, and it suggests that longer training on both
datasets can further improve the transfer learning performance of our models.
For simplicity, by default we do not replicate any dataset for training our models.

{\noindent \bf Sample replication.} We replicate samples for both images and videos, and in
each case measure a single epoch as training over the total samples, counting replicated samples, \ie
replication maintains the same number of training iterations. We train models with different
replication factors and show the final transfer accuracy and the normalized training time
in~\cref{fig:graphs_repeat_sampling_image,fig:graphs_repeat_sampling_video}. Sample replication
leads to faster training while maintaining or even improving the final transfer accuracy. Since a
large portion of the input sample is masked, replicating the sample multiple number of times
still provides enough learning signal for the model and does not
lead to a degradation in performance. This becomes even more relevant for \OURS's final settings
where we use higher masking ratios. For video data, we use an optimized dataloader (see details
in~\cref{appendix:impl_details}) with asynchronous I/O and fast video decoding. Even with this
optimized dataloader, sample replication leads to 20\% faster training.
We believe this is an important practical observation as video
dataloading is often a bottleneck for training models.

\subsection{Comparison to Prior Work}

In~\cref{tab:sota}, we compare \OURS's representation on our image and video classification benchmarks
with other \sota self-supervised methods. We focus on methods which use a Transformer backbone, ViT, or an
architecture based on Transformers like \swin.
To the best of our knowledge, prior work does not explore joint pre-training except for
BEVT~\cite{wang2021bevt}, which does masked image modeling followed by joint masked image and video modeling.
Unlike \OURS, BEVT only focuses on video representations and does not evaluate on image recognition.

\OURS is pretrained for 1600 epochs (\vitH for 2400 epochs).
Training with \imnetShort and \kineticsShort or \imnetShort and \sthsthShort leads to similar results, so we use the latter when scaling model size.
\OURS performs competitively for \emph{both} image
and video recognition when compared to the best models trained separately for either modality.
\OURS also performs favorably compared to methods like MaskedFeat which are pretrained only for video with specialized architectures on larger video datasets.
\OURS serves as a competitive initialization for transferring models on all modalities.
More notably, \OURS's performance improves significantly with larger architectures.
Using \vitH, \OURS \textbf{performs better or within the margin of error across all benchmarks}. Notably, \OURS even outperforms
concurrent works which use similar approaches while solely focusing on video representation learning.

\section{Conclusion and Future Work}
\label{sec:conclusion}

We present \OURS, a unified Transformer for images and videos that can be pretrained using masked autoencoding.
\OURS uses a simple architecture with minimal vision-specific changes and is competitive with specialized architectures and models tailored for images and videos.
We believe such generic models and approaches are a critical and under explored area of representation learning.
\OURS has not been empirically validated on other visual modalities like 3D, or non-visual modalities like audio.
The simplicity and generality of \OURS can likely enable future multi-modal systems that use shared parameters to model multiple modalities.
We only studied masked autoencoder based pretraining strategies for \OURS since they are easier to train.
However, pixel reconstruction methods do not learn good linearly separable features or similarity metrics~\cite{he2021masked}.
These properties are essential when finetuning with limited labeled data.
We believe that exploring other pretraining strategies, particularly in the context of joint modeling, will likely lead to improved models.
Similarly, joint finetuning strategies~\cite{girdhar2022omnivore} could be used to evaluate such unified models and lead to potentially even superior performance.

\paragraph{Ethical considerations.}
We study models for visual recognition on images and videos and our technical contributions are neutral from an ethics standpoint.
We do not propose a new dataset or make claims about the suitability of our model on production data.
Given their self-supervised nature, our learned representations are free from label bias, however they are still susceptible to the bias in the distribution of the visual data
We believe all ethical considerations that apply to visual recognition models equally apply to our model.

\bibliographystyle{ieee_fullname}
\bibliography{refs}

\begin{thebibliography}{10}\itemsep=-1pt

\bibitem{arandjelovic2017look}
Relja Arandjelovic and Andrew Zisserman.
\newblock Look, listen and learn.
\newblock In {\em ICCV}, 2017.

\bibitem{arnab2021vivit}
Anurag Arnab, Mostafa Dehghani, Georg Heigold, Chen Sun, Mario Lucic, and
  Cordelia Schmid.
\newblock {ViViT}: {A} video vision transformer.
\newblock In {\em ICCV}, 2021.

\bibitem{asano2019self}
Yuki~Markus Asano, Christian Rupprecht, and Andrea Vedaldi.
\newblock Self-labelling via simultaneous clustering and representation
  learning.
\newblock In {\em ICLR}, 2020.

\bibitem{atito2021sit}
Sara Atito, Muhammad Awais, and Josef Kittler.
\newblock Sit: Self-supervised vision transformer.
\newblock {\em arXiv preprint arXiv:2104.03602}, 2021.

\bibitem{baevski2022data2vec}
Alexei Baevski, Wei-Ning Hsu, Qiantong Xu, Arun Babu, Jiatao Gu, and Michael
  Auli.
\newblock data2vec: A general framework for self-supervised learning in speech,
  vision and language.
\newblock In {\em ICML}, 2022.

\bibitem{bao2021beit}
Hangbo Bao, Li Dong, and Furu Wei.
\newblock {BEiT}: {B}ert pre-training of image transformers.
\newblock In {\em ICLR}, 2022.

\bibitem{bardes2022vicreg}
Adrien Bardes, Jean Ponce, and Yann LeCun.
\newblock Vicreg: Variance-invariance-covariance regularization for
  self-supervised learning.
\newblock In {\em ICLR}, 2022.

\bibitem{berman2019multigrain}
Maxim Berman, Herv{\'e} J{\'e}gou, Andrea Vedaldi, Iasonas Kokkinos, and
  Matthijs Douze.
\newblock Multigrain: a unified image embedding for classes and instances.
\newblock {\em arXiv preprint arXiv:1902.05509}, 2019.

\bibitem{bertasius2021is}
Gedas Bertasius, Heng Wang, and Lorenzo Torresani.
\newblock Is space-time attention all you need for video understanding?
\newblock In {\em ICML}, 2021.

\bibitem{bourlard1988auto}
Herv{\'e} Bourlard and Yves Kamp.
\newblock Auto-association by multilayer perceptrons and singular value
  decomposition.
\newblock {\em Biological Cybernetics}, 1988.

\bibitem{caron2018deep}
Mathilde Caron, Piotr Bojanowski, Armand Joulin, and Matthijs Douze.
\newblock Deep clustering for unsupervised learning of visual features.
\newblock In {\em ECCV}, 2018.

\bibitem{caron2020unsupervised}
Mathilde Caron, Ishan Misra, Julien Mairal, Priya Goyal, Piotr Bojanowski, and
  Armand Joulin.
\newblock Unsupervised learning of visual features by contrasting cluster
  assignments.
\newblock In {\em NeurIPS}, 2020.

\bibitem{caron2021emerging}
Mathilde Caron, Hugo Touvron, Ishan Misra, Herv\'e J\'egou, Julien Mairal,
  Piotr Bojanowski, and Armand Joulin.
\newblock Emerging properties in self-supervised vision transformers.
\newblock In {\em ICCV}, 2021.

\bibitem{carreira2017quo}
Jo\~ao Carreira and Andrew Zisserman.
\newblock Quo vadis, action recognition? {A} new model and the kinetics
  dataset.
\newblock In {\em CVPR}, 2017.

\bibitem{castrejon2016learning}
Lluis Castrejon, Yusuf Aytar, Carl Vondrick, Hamed Pirsiavash, and Antonio
  Torralba.
\newblock Learning aligned cross-modal representations from weakly aligned
  data.
\newblock In {\em CVPR}, 2016.

\bibitem{chen2020generative}
Mark Chen, Alec Radford, Rewon Child, Jeffrey Wu, Heewoo Jun, David Luan, and
  Ilya Sutskever.
\newblock Generative pretraining from pixels.
\newblock In {\em ICML}, 2020.

\bibitem{chen2020simple}
Ting Chen, Simon Kornblith, Mohammad Norouzi, and Geoffrey Hinton.
\newblock A simple framework for contrastive learning of visual
  representations.
\newblock In {\em ICML}, 2020.

\bibitem{chen2020exploring}
Xinlei Chen and Kaiming He.
\newblock Exploring simple siamese representation learning.
\newblock In {\em CVPR}, 2021.

\bibitem{chen2021empirical}
Xinlei Chen, Saining Xie, and Kaiming He.
\newblock An empirical study of training self-supervised vision transformers.
\newblock In {\em ICCV}, 2021.

\bibitem{clark2020electra}
Kevin Clark, Minh-Thang Luong, Quoc~V Le, and Christopher~D Manning.
\newblock Electra: Pre-training text encoders as discriminators rather than
  generators.
\newblock {\em ICLR}, 2020.

\bibitem{cubuk2020randaugment}
Ekin~D Cubuk, Barret Zoph, Jonathon Shlens, and Quoc~V Le.
\newblock Randaugment: Practical automated data augmentation with a reduced
  search space.
\newblock In {\em CVPR}, 2020.

\bibitem{damen2021rescaling}
Dima Damen, Hazel Doughty, Giovanni~Maria Farinella, Antonino Furnari,
  Evangelos Kazakos, Jian Ma, Davide Moltisanti, Jonathan Munro, Toby Perrett,
  Will Price, et~al.
\newblock Rescaling egocentric vision.
\newblock {\em IJCV}, 2021.

\bibitem{devlin2018bert}
Jacob Devlin, Ming-Wei Chang, Kenton Lee, and Kristina Toutanova.
\newblock {BERT}: {P}re-training of deep bidirectional transformers for
  language understanding.
\newblock In {\em NAACL}, 2018.

\bibitem{dosovitskiy2020image}
Alexey Dosovitskiy, Lucas Beyer, Alexander Kolesnikov, Dirk Weissenborn,
  Xiaohua Zhai, Thomas Unterthiner, Mostafa Dehghani, Matthias Minderer, Georg
  Heigold, Sylvain Gelly, et~al.
\newblock An image is worth 16x16 words: Transformers for image recognition at
  scale.
\newblock In {\em ICLR}, 2021.

\bibitem{el2021large}
Alaaeldin El-Nouby, Gautier Izacard, Hugo Touvron, Ivan Laptev, Herv{\'e}
  Jegou, and Edouard Grave.
\newblock Are large-scale datasets necessary for self-supervised pre-training?
\newblock {\em arXiv preprint arXiv:2112.10740}, 2021.

\bibitem{ermolov2020whitening}
Aleksandr Ermolov, Aliaksandr Siarohin, Enver Sangineto, and Nicu Sebe.
\newblock Whitening for self-supervised representation learning.
\newblock In {\em ICML}, 2021.

\bibitem{fan2021multiscale}
Haoqi Fan, Bo Xiong, Karttikeya Mangalam, Yanghao Li, Zhicheng Yan, Jitendra
  Malik, and Christoph Feichtenhofer.
\newblock Multiscale vision transformers.
\newblock In {\em ICCV}, 2021.

\bibitem{feichtenhofer2022masked}
Christoph Feichtenhofer, Haoqi Fan, Yanghao Li, and Kaiming He.
\newblock Masked autoencoders as spatiotemporal learners.
\newblock {\em arXiv preprint arXiv:2205.09113}, 2022.

\bibitem{feichtenhofer2016convolutional}
Christoph Feichtenhofer, Axel Pinz, and Andrew Zisserman.
\newblock Convolutional two-stream network fusion for video action recognition.
\newblock In {\em CVPR}, 2016.

\bibitem{gallinari1987memoires}
Patrick Gallinari, Yann LeCun, Sylvie Thiria, and F~Fogelman Soulie.
\newblock M{\'e}moires associatives distribu{\'e}es: une comparaison
  (distributed associative memories: a comparison).
\newblock In {\em COGNITIVA}. 1987.

\bibitem{girdhar2019video}
Rohit Girdhar, Jo\~ao Carreira, Carl Doersch, and Andrew Zisserman.
\newblock Video action transformer network.
\newblock In {\em CVPR}, 2019.

\bibitem{girdhar2021anticipative}
Rohit Girdhar and Kristen Grauman.
\newblock {Anticipative Video Transformer}.
\newblock In {\em ICCV}, 2021.

\bibitem{girdhar2022omnivore}
Rohit Girdhar, Mannat Singh, Nikhila Ravi, Laurens van~der Maaten, Armand
  Joulin, and Ishan Misra.
\newblock {Omnivore: A Single Model for Many Visual Modalities}.
\newblock In {\em CVPR}, 2022.

\bibitem{gong2014improving}
Yunchao Gong, Liwei Wang, Micah Hodosh, Julia Hockenmaier, and Svetlana
  Lazebnik.
\newblock Improving image-sentence embeddings using large weakly annotated
  photo collections.
\newblock In {\em ECCV}, 2014.

\bibitem{goyal2022vision}
Priya Goyal, Quentin Duval, Isaac Seessel, Mathilde Caron, Mannat Singh, Ishan
  Misra, Levent Sagun, Armand Joulin, and Piotr Bojanowski.
\newblock Vision models are more robust and fair when pretrained on uncurated
  images without supervision.
\newblock {\em arXiv preprint arXiv:2202.08360}, 2022.

\bibitem{goyal2019scaling}
Priya Goyal, Dhruv Mahajan, Abhinav Gupta, and Ishan Misra.
\newblock Scaling and benchmarking self-supervised visual representation
  learning.
\newblock In {\em ICCV}, 2019.

\bibitem{goyal2017something}
Raghav Goyal, Samira Ebrahimi~Kahou, Vincent Michalski, Joanna Materzynska,
  Susanne Westphal, Heuna Kim, Valentin Haenel, Ingo Fruend, Peter Yianilos,
  Moritz Mueller-Freitag, Florian Hoppe, Christian Thurau, Ingo Bax, and Roland
  Memisevic.
\newblock The ``something something'' video database for learning and
  evaluating visual common sense.
\newblock In {\em ICCV}, 2017.

\bibitem{grill2020bootstrap}
Jean-Bastien Grill, Florian Strub, Florent Altch{\'e}, Corentin Tallec, Pierre
  Richemond, Elena Buchatskaya, Carl Doersch, Bernardo Avila~Pires, Zhaohan
  Guo, Mohammad Gheshlaghi~Azar, et~al.
\newblock Bootstrap your own latent-a new approach to self-supervised learning.
\newblock {\em NeurIPS}, 2020.

\bibitem{hadsell2006dimensionality}
Raia Hadsell, Sumit Chopra, and Yann LeCun.
\newblock Dimensionality reduction by learning an invariant mapping.
\newblock In {\em CVPR}, 2006.

\bibitem{he2021masked}
Kaiming He, Xinlei Chen, Saining Xie, Yanghao Li, Piotr Doll{\'a}r, and Ross
  Girshick.
\newblock Masked autoencoders are scalable vision learners.
\newblock In {\em CVPR}, 2022.

\bibitem{he2020momentum}
Kaiming He, Haoqi Fan, Yuxin Wu, Saining Xie, and Ross Girshick.
\newblock Momentum contrast for unsupervised visual representation learning.
\newblock In {\em CVPR}, 2020.

\bibitem{hinton1993autoencoders}
Geoffrey~E Hinton and Richard Zemel.
\newblock Autoencoders, minimum description length and helmholtz free energy.
\newblock In {\em NeurIPS}, 1993.

\bibitem{hoffer2020augment}
Elad Hoffer, Tal Ben-Nun, Itay Hubara, Niv Giladi, Torsten Hoefler, and Daniel
  Soudry.
\newblock Augment your batch: Improving generalization through instance
  repetition.
\newblock In {\em CVPR}, 2020.

\bibitem{iNaturalist}
Grant~Van Horn, Oisin~Mac Aodha, Yang Song, Yin Cui, Chen Sun, Alex Shepard,
  Hartwig Adam, Pietro Perona, and Serge Belongie.
\newblock The inaturalist species classification and detection dataset.
\newblock In {\em CVPR}, 2018.

\bibitem{huang2016deep}
Gao Huang, Yu Sun, Zhuang Liu, Daniel Sedra, and Kilian~Q Weinberger.
\newblock Deep networks with stochastic depth.
\newblock In {\em ECCV}, 2016.

\bibitem{ioffe2015batch}
Sergey Ioffe and Christian Szegedy.
\newblock Batch normalization: Accelerating deep network training by reducing
  internal covariate shift.
\newblock In {\em ICML}, 2015.

\bibitem{karpathy2015deep}
Andrej Karpathy and Li Fei-Fei.
\newblock Deep visual-semantic alignments for generating image descriptions.
\newblock In {\em CVPR}, 2015.

\bibitem{kay2017kinetics}
Will Kay, Joao Carreira, Karen Simonyan, Brian Zhang, Chloe Hillier, Sudheendra
  Vijayanarasimhan, Fabio Viola, Tim Green, Trevor Back, Paul Natsev, AMustafa
  Suleyman, and Andrew Zisserman.
\newblock The kinetics human action video dataset.
\newblock {\em arXiv preprint arXiv:1705.06950}, 2017.

\bibitem{kramer1991nonlinear}
Mark~A Kramer.
\newblock Nonlinear principal component analysis using autoassociative neural
  networks.
\newblock {\em AIChE}, 1991.

\bibitem{le1987modeles}
Yann LeCun and Fran{\c{c}}oise Fogelman-Souli{\'e}.
\newblock Mod{\`e}les connexionnistes de l'apprentissage.
\newblock {\em Intellectica}, 1987.

\bibitem{junnan2021prototypical}
Junnan Li, Pan Zhou, Caiming Xiong, and Steven~C.H. Hoi.
\newblock Prototypical contrastive learning of unsupervised representations.
\newblock In {\em ICLR}, 2021.

\bibitem{li2022improved}
Yanghao Li, Chao-Yuan Wu, Haoqi Fan, Karttikeya Mangalam, Bo Xiong, Jitendra
  Malik, and Christoph Feichtenhofer.
\newblock Improved multiscale vision transformers for classification and
  detection.
\newblock In {\em CVPR}, 2022.

\bibitem{anonymous2021polyvit}
Valerii Likhosherstov, Anurag Arnab, Krzysztof Choromanski, Mario Lucic, Yi
  Tay, Adrian Weller, and Mostafa Dehghani.
\newblock Polyvit: Co-training vision transformers on images, videos and audio.
\newblock {\em arXiv preprint arXiv:2111.12993}, 2021.

\bibitem{liu2022swinv2}
Ze Liu, Han Hu, Yutong Lin, Zhuliang Yao, Zhenda Xie, Yixuan Wei, Jia Ning, Yue
  Cao, Zheng Zhang, Li Dong, Furu Wei, and Baining Guo.
\newblock Swin transformer v2: Scaling up capacity and resolution.
\newblock In {\em CVPR}, 2022.

\bibitem{liu2021swin}
Ze Liu, Yutong Lin, Yue Cao, Han Hu, Yixuan Wei, Zheng Zhang, Stephen Lin, and
  Baining Guo.
\newblock Swin transformer: Hierarchical vision transformer using shifted
  windows.
\newblock In {\em ICCV}, 2021.

\bibitem{liu2021video}
Ze Liu, Jia Ning, Yue Cao, Yixuan Wei, Zheng Zhang, Stephen Lin, and Han Hu.
\newblock Video swin transformer.
\newblock {\em arXiv preprint arXiv:2106.13230}, 2021.

\bibitem{liu2021group}
Ze Liu, Zheng Zhang, Yue Cao, Han Hu, and Xin Tong.
\newblock Group-free 3d object detection via transformers.
\newblock In {\em ICCV}, 2021.

\bibitem{lu202012}
Jiasen Lu, Vedanuj Goswami, Marcus Rohrbach, Devi Parikh, and Stefan Lee.
\newblock 12-in-1: Multi-task vision and language representation learning.
\newblock In {\em CVPR}, 2020.

\bibitem{miech2020end}
Antoine Miech, Jean-Baptiste Alayrac, Lucas Smaira, Ivan Laptev, Josef Sivic,
  and Andrew Zisserman.
\newblock End-to-end learning of visual representations from uncurated
  instructional videos.
\newblock In {\em CVPR}, 2020.

\bibitem{misra2021end}
Ishan Misra, Rohit Girdhar, and Armand Joulin.
\newblock {An End-to-End Transformer Model for 3D Object Detection}.
\newblock In {\em ICCV}, 2021.

\bibitem{misra2020self}
Ishan Misra and Laurens van~der Maaten.
\newblock Self-supervised learning of pretext-invariant representations.
\newblock In {\em CVPR}, 2020.

\bibitem{morgado2021robust}
Pedro Morgado, Ishan Misra, and Nuno Vasconcelos.
\newblock Robust audio-visual instance discrimination.
\newblock In {\em CVPR}, 2021.

\bibitem{morgado2021audio}
Pedro Morgado, Nuno Vasconcelos, and Ishan Misra.
\newblock Audio-visual instance discrimination with cross-modal agreement.
\newblock In {\em CVPR}, 2021.

\bibitem{olshausen1996}
B.~A. Olshausen and D.~J. Field.
\newblock Emergence of simple-cell receptive field properties by learning a
  sparse code for natural images.
\newblock {\em Nature}, 1996.

\bibitem{oord2018representation}
Aaron van~den Oord, Yazhe Li, and Oriol Vinyals.
\newblock Representation learning with contrastive predictive coding.
\newblock In {\em NeurIPS}, 2018.

\bibitem{owens}
Andrew Owens and Alexei~A Efros.
\newblock Audio-visual scene analysis with self-supervised multisensory
  features.
\newblock In {\em ECCV}, 2018.

\bibitem{parmar2018image}
Niki Parmar, Ashish Vaswani, Jakob Uszkoreit, Lukasz Kaiser, Noam Shazeer,
  Alexander Ku, and Dustin Tran.
\newblock Image transformer.
\newblock In {\em ICML}, 2018.

\bibitem{pathak2016context}
Deepak Pathak, Philipp Krahenbuhl, Jeff Donahue, Trevor Darrell, and Alexei~A
  Efros.
\newblock Context encoders: Feature learning by inpainting.
\newblock In {\em CVPR}, 2016.

\bibitem{polyak1992acceleration}
Boris~T Polyak and Anatoli~B Juditsky.
\newblock Acceleration of stochastic approximation by averaging.
\newblock {\em SIAM journal on control and optimization}, 1992.

\bibitem{ILSVRC15}
Olga Russakovsky, Jia Deng, Hao Su, Jonathan Krause, Sanjeev Satheesh, Sean Ma,
  Zhiheng Huang, Andrej Karpathy, Aditya Khosla, Michael Bernstein,
  Alexander~C. Berg, and Li Fei-Fei.
\newblock {ImageNet Large Scale Visual Recognition Challenge}.
\newblock {\em IJCV}, 2015.

\bibitem{salakhutdinov2009deep}
Ruslan Salakhutdinov and Geoffrey Hinton.
\newblock Deep {B}oltzmann machines.
\newblock In {\em AISTATS}, 2009.

\bibitem{simonyan2014two}
Karen Simonyan and Andrew Zisserman.
\newblock Two-stream convolutional networks for action recognition in videos.
\newblock In {\em NeurIPS}, 2014.

\bibitem{szegedy2016rethinking}
Christian Szegedy, Vincent Vanhoucke, Sergey Ioffe, Jon Shlens, and Zbigniew
  Wojna.
\newblock Rethinking the inception architecture for computer vision.
\newblock In {\em CVPR}, 2016.

\bibitem{tan2021vimpac}
Hao Tan, Jie Lei, Thomas Wolf, and Mohit Bansal.
\newblock {VIMPAC}: {V}ideo pre-training via masked token prediction and
  contrastive learning.
\newblock {\em arXiv preprint arXiv:2106.11250}, 2021.

\bibitem{tian2019contrastive}
Yonglong Tian, Dilip Krishnan, and Phillip Isola.
\newblock Contrastive multiview coding.
\newblock {\em arXiv preprint arXiv:1906.05849}, 2019.

\bibitem{tong2022videomae}
Zhan Tong, Yibing Song, Jue Wang, and Limin Wang.
\newblock Videomae: Masked autoencoders are data-efficient learners for
  self-supervised video pre-training.
\newblock {\em arXiv preprint arXiv:2203.12602}, 2022.

\bibitem{touvron2021training}
Hugo Touvron, Matthieu Cord, Matthijs Douze, Francisco Massa, Alexandre
  Sablayrolles, and Herv{\'e} J{\'e}gou.
\newblock Training data-efficient image transformers \& distillation through
  attention.
\newblock In {\em ICML}, 2021.

\bibitem{vaswani2017attention}
Ashish Vaswani, Noam Shazeer, Niki Parmar, Jakob Uszkoreit, Llion Jones,
  Aidan~N Gomez, Lukasz Kaiser, and Illia Polosukhin.
\newblock Attention is all you need.
\newblock In {\em NeurIPS}, 2017.

\bibitem{vincent2008extracting}
P. Vincent, H. Larochelle, Y. Bengio, and P.-A. Manzagol.
\newblock Extracting and composing robust features with denoising autoencoders.
\newblock In {\em ICML}, 2008.

\bibitem{wang2021bevt}
Rui Wang, Dongdong Chen, Zuxuan Wu, Yinpeng Chen, Xiyang Dai, Mengchen Liu,
  Yu-Gang Jiang, Luowei Zhou, and Lu Yuan.
\newblock {BEVT}: {B}ert pretraining of video transformers.
\newblock In {\em CVPR}, 2022.

\bibitem{wang2021pyramid}
Wenhai Wang, Enze Xie, Xiang Li, Deng-Ping Fan, Kaitao Song, Ding Liang, Tong
  Lu, Ping Luo, and Ling Shao.
\newblock Pyramid vision transformer: A versatile backbone for dense prediction
  without convolutions.
\newblock In {\em ICCV}, 2021.

\bibitem{wei2021masked}
Chen Wei, Haoqi Fan, Saining Xie, Chao-Yuan Wu, Alan Yuille, and Christoph
  Feichtenhofer.
\newblock Masked feature prediction for self-supervised visual pre-training.
\newblock In {\em CVPR}, 2022.

\bibitem{wu2018unsupervised}
Zhirong Wu, Yuanjun Xiong, Stella~X Yu, and Dahua Lin.
\newblock Unsupervised feature learning via non-parametric instance
  discrimination.
\newblock In {\em CVPR}, 2018.

\bibitem{xie2021simmim}
Zhenda Xie, Zheng Zhang, Yue Cao, Yutong Lin, Jianmin Bao, Zhuliang Yao, Qi
  Dai, and Han Hu.
\newblock Simmim: A simple framework for masked image modeling.
\newblock In {\em CVPR}, 2022.

\bibitem{yan2020cluster}
Xueting Yan, Ishan Misra, Abhinav Gupta, Deepti Ghadiyaram, and Dhruv Mahajan.
\newblock {ClusterFit: Improving Generalization of Visual Representations}.
\newblock In {\em CVPR}, 2020.

\bibitem{yun2019cutmix}
Sangdoo Yun, Dongyoon Han, Seong~Joon Oh, Sanghyuk Chun, Junsuk Choe, and
  Youngjoon Yoo.
\newblock Cutmix: Regularization strategy to train strong classifiers with
  localizable features.
\newblock In {\em ICCV}, 2019.

\bibitem{zbontar2021barlow}
Jure Zbontar, Li Jing, Ishan Misra, Yann LeCun, and St{\'e}phane Deny.
\newblock Barlow twins: Self-supervised learning via redundancy reduction.
\newblock In {\em ICML}, 2021.

\bibitem{zhang2017mixup}
Hongyi Zhang, Moustapha Cisse, Yann~N Dauphin, and David Lopez-Paz.
\newblock mixup: Beyond empirical risk minimization.
\newblock In {\em ICLR}, 2018.

\bibitem{zhao2020point}
Hengshuang Zhao, Li Jiang, Jiaya Jia, Philip Torr, and Vladlen Koltun.
\newblock Point transformer.
\newblock In {\em ICCV}, 2021.

\bibitem{zhong2020random}
Zhun Zhong, Liang Zheng, Guoliang Kang, Shaozi Li, and Yi Yang.
\newblock Random erasing data augmentation.
\newblock In {\em AAAI}, 2020.

\bibitem{Places205}
Bolei Zhou, Agata Lapedriza, Jianxiong Xiao, Antonio Torralba, and Aude Oliva.
\newblock Learning deep features for scene recognition using places database.
\newblock In {\em NeurIPS}, 2014.

\bibitem{zhou2021ibot}
Jinghao Zhou, Chen Wei, Huiyu Wang, Wei Shen, Cihang Xie, Alan Yuille, and Tao
  Kong.
\newblock ibot: Image bert pre-training with online tokenizer.
\newblock In {\em ICLR}, 2022.

\end{thebibliography}

\clearpage
\appendix
\section{Datasets}\label{appendix:datasets}

\subsection{Image datasets}
\par \noindent \textbf{\imnet (\imnetShort)~\cite{ILSVRC15}.} We use the ILSVRC 2012 challenge subset of
ImageNet that has 1.28M training and 50K val images with 1000 classes. The 1000 classes cover a wide
range of concepts from fine-grained species of dogs, to every day indoor and outdoor objects. The
dataset is released under a non-commercial license. This subset of ImageNet is widely used for
benchmarking image recognition models.

\par \noindent \textbf{\inat (\inatShort)~\cite{iNaturalist}.} The iNaturalist dataset is a fine-grained plant
and animal species classification dataset. We use the 2018 version of the dataset that has 437K
training and 24K val images with 8142 classes. The dataset was collected in collaboration with
iNaturalist, a citizen science effort that uses pictures submitted by people around the world. The
dataset is released under a non-commercial iNaturalist license. To the best of our knowledge, no PII
or harmful content has been reported in the dataset.

\par \noindent \textbf{\placesThree (\placesThreeShort)~\cite{Places205}.} The Places dataset is a scene
recognition dataset that evaluates image recognition models on indoor and outdoor scene
classification. The dataset consists of 1.8M training and 36K val images with 365 scene classes. The
dataset has public images and is released under a non-commercial license. To the best of our
knowledge, no PII or harmful content has been reported in the dataset.

\subsection{Video datasets}

\par \noindent \textbf{\sthsth (\sthsthShort)~\cite{goyal2017something}.} This is a video action classification
dataset with a special emphasis on temporal modeling. It consists of $\sim 169K$ training and $\sim
25K$ validation clips, each a few seconds long, classified into one of 174 action classes. Due to
the nature of the classes considered (\eg ``covering something'' and ``uncovering something''), the
dataset requires temporal modeling to correctly classify each video. The dataset has been collected
by consenting participants who recorded the videos given the action label, and released under a
non-commercial license. To the best of our knowledge, no PII or harmful content has been reported in
the dataset.

\par \noindent \textbf{\epic (\epicShort)~\cite{damen2021rescaling}.}
This dataset consists of 100 hours total of unscripted egocentric videos. Each video is densely
labeled with human-object interactions (``clips''), which consists of a start time, end time, one of
300 nouns (the object interacted with) and one of 97 verbs (the type of interaction). There are
$\sim 67K$  %
training and $\sim 10K$  %
validation clips. Following prior work~\cite{girdhar2022omnivore,damen2021rescaling}, we tackle the
task of recognizing the 3,806 {\tt (verb, noun)} pairs given a clip. Note that not all {\tt (verb,
noun)} combinations occur in both training and testing data. We use this dataset as a transfer task
to evaluate the learned representation. The data is released under CC-BY-NC 4.0 license. The videos
were collected by consenting participants who wore egocentric cameras while recording their daily
activities, typically cooking. Given the egocentric nature of the videos, PII such as faces are not
visible in the videos. To the best of our knowledge, no offensive content has been reported on this
dataset.

\par \noindent \textbf{\kinetics (\kineticsShort)~\cite{kay2017kinetics}.} This dataset consists of $\sim 240K$
training and $\sim 20K$ validation third-person video clips that are 10 seconds in length. Each clip
is labeled into one of 400 action categories. The task requires classifying each validation video
into one of these categories. The dataset is based on publicly available web videos from YouTube.
Due to the videos being taken down over time, the dataset changes over time making apples-to-apples
comparison with prior work difficult. Hence, we use a static dataset like \sthsthShort for
pre-training and the primary comparisons. We will release the set of train and test videos that we
had access to from this dataset. To the best of our knowledge, no PII or harmful content has been
reported on this dataset.

\section{Implementation Details}
\label{appendix:impl_details}

\subsection{Details about Pretraining}
\label{appendix:pretrain_details}

We pretrain the model jointly on \imnetShort and \sthsthShort using the hyperparameters in \cref{tab:pretrain_settings}.
The dataset-specific hyperparameters are in the individual columns, and others are in the middle.
These apply to \vitB, \vitL, and \vitH unless specified otherwise.
We use the same hyperparameters for pretraining the models on \imnetShort and \kineticsShort (\cref{tab:sota}).
The \vitH model in~\cref{tab:sota} is pretrained for 2400 epochs.

\begin{table}[!htb]
\begin{center}
    \centering
    \begin{tabular}{l|cc}
        Config & \imnetShort & \sthsthShort \\
        \midrule
        Optimizer & \multicolumn{2}{c}{AdamW} \\
        Peak learning rate & \multicolumn{2}{c}{3e-4} \\
        Weight decay & \multicolumn{2}{c}{0.05} \\
        Optimizer Momentum & \multicolumn{2}{c}{$\beta_1=0.9,
        \beta_2=0.95$~\cite{chen2020generative}} \\
        Batch size & \multicolumn{2}{c}{2048} \\
        Sample replication & 1 & 4 \\
        Warmup epochs & \multicolumn{2}{c}{40} \\
        Total epochs & \multicolumn{2}{c}{800 (default), 1600 (\cref{tab:sota})} \\
        Augmentations: \\
        \quad {\tt ShortSideScale} & N/A & 256px \\
        \quad {\tt RandomResizedCrop} \\
        \qquad {\tt size} & \multicolumn{2}{c}{224px} \\
        \qquad {\tt scale} & [0.2, 1.0] & [0.08, 1.0] \\
        \qquad {\tt ratio} & \multicolumn{2}{c}{[0.75, 1.33]} \\
        \qquad {\tt interpolation} & Bicubic & Bilinear \\
        \quad {\tt RandomHorizontalFlip} & $p=0.5$ & $p=0.0$ \\
        \quad {\tt Normalize} & \multicolumn{2}{c}{Yes}
    \end{tabular}
\end{center}
\caption{Pretraining hyperparameters}
\label{tab:pretrain_settings}
\end{table}

We train the model using 64 (or 128 for \vitL, \vitH) 32GB+ GPUs (A100 or Volta 32GB). The model is trained
with an $\ell_{2}$ loss on the pixel values. We normalize the target using the mean and variance of
the pixels in the patch, where the norm is computed for each color channel separately, before
applying the loss. Note that we use the original 0-255 pixel values before normalizing for this
loss. We use a a decoder with 4 layers and 384 dimension for \vitB, 4 layers and 512 dimension for \vitL, and 8 layers and 512 dimension for \vitH.

\par \noindent \textbf{Optimized video dataloading.}
AI training clusters
usually load data from high-latency high-throughput filesystems, where
dataloading for image and video datasets is bottlenecked not by the throughput, but by the
latency for each read.
For videos, this problem is exacerbated by the additional time spent decoding videos,
ultimately resulting in situations where video training is bottlenecked by dataloading.
In \OURS, where the training step is lightweight, owing to 95\% masking for videos,
in order to see a congruent improvement in wall-clock times, we needed to optimize our
video dataloading.
PyTorch dataloaders read data in synchronous fashion, which means that while a sample is
loaded from disk and decoded, the corresponding dataloading process blocks without doing
any work while waiting.
Having multiple dataloader workers can mitigate this, but there is a cap to it
based on the CPU memory and cores. To mitigate this issue, we implemented asynchronous
dataloaders using \texttt{asyncio}, allowing the
same process to send multiple requests without blocking. This allows us to reduce the
amortized data reading time to effectively zero, resulting in much faster training speeds.
As~\cref{fig:graphs_repeat_sampling_video} shows, our sample replication strategy provides a training speedup over this optimized dataloading setup.

\subsection{Transfer Learning}
\label{appendix:transfer_details}

\cref{tab:im_ft_settings} specifies the hyperparameters for finetuning \vitB, \vitL and \vitH
on the image datasets we utilize -- \imnetShort, \inatShort and \placesThreeShort.
We report the peak test accuracy observed during training.
For all three datasets we use the same settings with just different peak learning rates
and total epochs.

\begin{table}[t]
\begin{center}
    \centering
    \begin{tabular}{l|cc}
        Config & \vitB & ViT-\{L, H\} \\
        \midrule
        Optimizer & \multicolumn{2}{c}{AdamW} \\
        Peak learning rate \\
        \quad {\imnetShort} & 4e-3 & 2e-3 \\
        \quad {\inatShort} & \multicolumn{2}{c}{2e-3} \\
        \quad {\placesThreeShort} & \multicolumn{2}{c}{2e-3} \\
        Total epochs \\
        \quad {\imnetShort} & 100 & 50 \\
        \quad {\inatShort} & 300 & 100 \\
        \quad {\placesThreeShort} & 60 & 50 \\
        Warmup epochs & \multicolumn{2}{c}{5} \\
        Weight decay & 1e-4 & 5e-2 \\
        Layerwise LR decay~\cite{clark2020electra, bao2021beit} & 0.65 & 0.75 \\
        Optimizer Momentum & \multicolumn{2}{c}{$\beta_1=0.9,
        \beta_2=0.999$} \\
        Batch size & \multicolumn{2}{c}{1024} \\
        DropPath~\cite{huang2016deep} & 0.1 & 0.2 \\
        EMA~\cite{polyak1992acceleration} &  \multicolumn{2}{c}{1e-4} \\
        Augmentations: \\
        \quad {\tt RandomResizedCrop} \\
        \qquad {\tt size} & \multicolumn{2}{c}{224px} \\
        \qquad {\tt scale} & \multicolumn{2}{c}{[0.08, 1.0]} \\
        \qquad {\tt ratio} & \multicolumn{2}{c}{[0.75, 1.33]} \\
        \qquad {\tt interpolation} & \multicolumn{2}{c}{Bicubic} \\
        \quad {\tt RandomHorizontalFlip} &\multicolumn{2}{c}{$p=0.5$} \\
        \quad {\tt RandomAugment~\cite{cubuk2020randaugment}} \\
        \qquad {\tt magnitude} & \multicolumn{2}{c}{9} \\
        \qquad {\tt num\_layers} & \multicolumn{2}{c}{0.5} \\
        \quad {\tt RandomErasing~\cite{zhong2020random}} & \multicolumn{2}{c}{$p=0.25$} \\
        \quad {\tt Normalize} & \multicolumn{2}{c}{Yes} \\
        \quad {\tt mixup~\cite{zhang2017mixup}} & \multicolumn{2}{c}{0.8} \\
        \quad {\tt CutMix~\cite{yun2019cutmix}} & \multicolumn{2}{c}{1.0} \\
        \quad {\tt LabelSmoothing~\cite{szegedy2016rethinking}} & \multicolumn{2}{c}{0.1} \\
    \end{tabular}
\end{center}
\caption{Finetuning hyperparameters for \imnetShort, \inatShort and \placesThreeShort}
\label{tab:im_ft_settings}
\end{table}

For fine tuning on video datasets, we sample 16 frames from clips. For \sthsthShort
and \epicShort we sample 2.7 second clips. For \kineticsShort we sample 2 second clips.
At test time, we sample 5 clips with 3 spatial crops and report the final test accuracy at the end
of training. \cref{tab:vid_ft_settings} specifies the hyperparameters for finetuning on \sthsthShort and
\epicShort, and~\cref{tab:vid_ft_settings_k400} on \kineticsShort.

For all the datasets, we use the same finetuning hyperparameters for \vitL and \vitH.

\begin{table}
    \centering
    \begin{tabular}[t]{l|cc}
        Config & \vitB & ViT-\{L, H\} \\
        \midrule
        Optimizer & \multicolumn{2}{c}{AdamW} \\
        Peak learning rate & \multicolumn{2}{c}{1e-3} \\
        Total epochs & \multicolumn{2}{c}{40} \\
        Batch size & \multicolumn{2}{c}{512} \\
        Warmup epochs & \multicolumn{2}{c}{5} \\
        Weight decay & \multicolumn{2}{c}{5e-2} \\
        Layerwise LR decay~\cite{clark2020electra, bao2021beit} & \multicolumn{2}{c}{0.75} \\
        Optimizer Momentum & \multicolumn{2}{c}{$\beta_1=0.9$} \\
         & \multicolumn{2}{c}{$\beta_2=0.999$} \\
        DropPath~\cite{huang2016deep} & 0.1 & 0.2 \\
        Augmentations: \\
        \quad {\tt ShortSideScale} & \multicolumn{2}{c}{256px} \\
        \quad {\tt RandomResizedCrop} \\
        \qquad {\tt size} & \multicolumn{2}{c}{224px} \\
        \qquad {\tt scale} & \multicolumn{2}{c}{[0.08, 1.0]} \\
        \qquad {\tt ratio} & \multicolumn{2}{c}{[0.75, 1.33]} \\
        \qquad {\tt interpolation} & \multicolumn{2}{c}{Bicubic} \\
        \quad {\tt RandomAugment~\cite{cubuk2020randaugment}} \\
        \qquad {\tt magnitude} & \multicolumn{2}{c}{7} \\
        \qquad {\tt num\_layers} & \multicolumn{2}{c}{4} \\
        \quad {\tt RandomErasing~\cite{zhong2020random}} & \multicolumn{2}{c}{$p=0.25$} \\
        \quad {\tt Normalize} & \multicolumn{2}{c}{Yes} \\
        \quad {\tt mixup~\cite{zhang2017mixup}} & \multicolumn{2}{c}{0.8} \\
        \quad {\tt CutMix~\cite{yun2019cutmix}} & \multicolumn{2}{c}{1.0} \\
        \quad {\tt LabelSmoothing~\cite{szegedy2016rethinking}} & \multicolumn{2}{c}{0.1} \\
    \end{tabular}
    \caption{Finetuning hyperparameters for \sthsthShort \& \epicShort.}
    \label{tab:vid_ft_settings}
\end{table}

\begin{table}
    \centering
    \begin{tabular}[t]{l|cc}
        Config & \vitB & ViT-\{L, H\} \\
        \midrule
        Optimizer & \multicolumn{2}{c}{AdamW} \\
        Peak learning rate & \multicolumn{2}{c}{1e-3} \\
        Total epochs & 175 & 100 \\
        Sample replication & \multicolumn{2}{c}{2} \\
        Batch size & 1024 & 512 \\
        Warmup epochs & \multicolumn{2}{c}{9} \\
        Weight decay & \multicolumn{2}{c}{5e-2} \\
        Layerwise LR decay~\cite{clark2020electra, bao2021beit} & \multicolumn{2}{c}{0.75} \\
        Optimizer Momentum & \multicolumn{2}{c}{$\beta_1=0.9$} \\
         & \multicolumn{2}{c}{$\beta_2=0.999$} \\
        DropPath~\cite{huang2016deep} & 0.1 & 0.2 \\
        Augmentations: \\
        \quad {\tt ShortSideScale} & \multicolumn{2}{c}{256px} \\
        \quad {\tt RandomResizedCrop} \\
        \qquad {\tt size} & \multicolumn{2}{c}{224px} \\
        \qquad {\tt scale} & \multicolumn{2}{c}{[0.08, 1.0]} \\
        \qquad {\tt ratio} & \multicolumn{2}{c}{[0.75, 1.33]} \\
        \qquad {\tt interpolation} & \multicolumn{2}{c}{Bicubic} \\
        \quad {\tt RandomHorizontalFlip} &\multicolumn{2}{c}{$p=0.5$} \\
        \quad {\tt RandomAugment~\cite{cubuk2020randaugment}} \\
        \qquad {\tt magnitude} & \multicolumn{2}{c}{9} \\
        \qquad {\tt num\_layers} & \multicolumn{2}{c}{2} \\
        \quad {\tt Normalize} & \multicolumn{2}{c}{Yes} \\
        \quad {\tt mixup~\cite{zhang2017mixup}} & \multicolumn{2}{c}{0.8} \\
        \quad {\tt CutMix~\cite{yun2019cutmix}} & \multicolumn{2}{c}{1.0} \\
        \quad {\tt LabelSmoothing~\cite{szegedy2016rethinking}} & \multicolumn{2}{c}{0.1} \\
    \end{tabular}
    \caption{Finetuning hyperparameters for \kineticsShort.}\label{tab:vid_ft_settings_k400}
\end{table}

\subsection{Details about Ablations}
\label{appendix:ablation_details}

For ablations, we start from a base pretraining configuration on \imnetShort and \sthsthShort
that includes 1) 75\% and 90\% masking on \imnetShort and \sthsthShort respectively; 2) Random and
Tube masking on \imnetShort and \sthsthShort respectively; 3) A common 4-layer 384D decoder for both
datasets; 4) Peak learning rate of 3e-4.

\par \noindent \textbf{Masking ratio.}
For the masking ratio ablation, we vary the amount of patches that are masked for each of the
modalities. We start from the default 75\% masking for each modality~\cite{he2021masked}, and
increase it upto 90\% for images and 95\% for videos. We notice that while downstream performance on
images is stable across different masking ratios on the image dataset during pretraining, increasing
the  video masking leads to improved performance on downstream video tasks. We observe the best
performance at 95\% masking on videos.

\par \noindent \textbf{Masking type.}
For this ablation, we vary the type of masking used in each modality. Starting from the default
masking type of (Random, Tube), we experiment with Causal masking on images, and Frame, Causal or
Random masking on videos. In all cases we keep the masking ratio to be 75\% for images and 90\% for
videos.

\par \noindent \textbf{Decoder capacity.} For this ablation, we explored different decoder capacities by varying
the decoder depth and embedding dimension. In particular, we test decoder depths of 2, 4 and 8
layers as well as embedding dimensions of 384 and 512.

\par \noindent \textbf{Sample replication.}
We briefly note the procedure for sample replication. Let $B$ denote the total batch size for
training without any replication. To maintain the total batch size for training, when replicating a
sample $t$ times, we sample $\frac{B}{t}$ training samples from the dataset. After replication, each
of the $B$ samples is augmented and processed individually. Thus, sample replication reduces the I/O
associated with reading and decoding a sample by a factor proportional to the replication factor
$t$.

\par \noindent \textbf{Dataset ratio.} We experiment with varying the relative dataset ratio of \imnetShort and
 \sthsthShort such that we replicate only one dataset at a time. In addition to the default dataset
 ratio of 1:1 (\imnetShort:\sthsthShort), we test dataset ratios of 1:2, 1:3, 2:1 and 3:1. For such
 dataset ratios, the samples for one of the datasets are replicated for every epoch, leading to
 longer training wall clock time.

\par \noindent \textbf{Specify random variance.}
Due to the large number of experiments and compute associated with training the models, we note the
random variance across a small subset of our experiments from~\cref{sec:ablations}. We measure the
random variance of both pretraining and transfer learning. We pretrain the model with different
random seeds and finetune it on \imnet and \sthsthShort. Across a trial of 3 such pretraining and 2
finetuning (total 6 runs), we observed a variance of $0.3$\% and $0.7$\% on \imnet and \sthsthShort
respectively.

\subsection{Visualization details}

To visualize the pixel reconstructions, we train another model without the patch mean/var
normalization in the loss. This ensures the model can directly generate the pixel values that we can
visualize without needing to provide it the patch's mean/variance. For visualization, we reshape the
predicted pixel values to the original image dimensions, and replace the unmasked patches with the
ground truth pixel values.

\section{Additional Visualizations}
We present additional visualizations in~\cref{fig:add_visualizations_video,fig:add_visualizations_img}.
To match the model's training settings, we visualize by masking 90\% of the image patches and 95\% of the video patches.
\begin{table}[t]
     \centering
     \setlength{\tabcolsep}{2pt}
     \newcommand{\raiseboxht}[0]{-0.45}
     \newcommand{\raiseboxlastht}[0]{\raiseboxht}
     \newcommand{\vizcropht}[0]{8.1cm}
     \newcommand{\vizcroplastht}[0]{16.3cm}
     \newcommand{\percolwidth}[0]{0.48}
     \newcommand{\perimgcolwidth}[0]{0.06}  %
     \hspace{-0.15in}  %
     \resizebox{\linewidth}{!}{
     \begin{tabular}{ccc}
         \multicolumn{2}{c}{ \color{VideoDark} \bf\sthsthShort}\\
         \includegraphics[width=\percolwidth\linewidth, trim={0 0 0 0},clip]{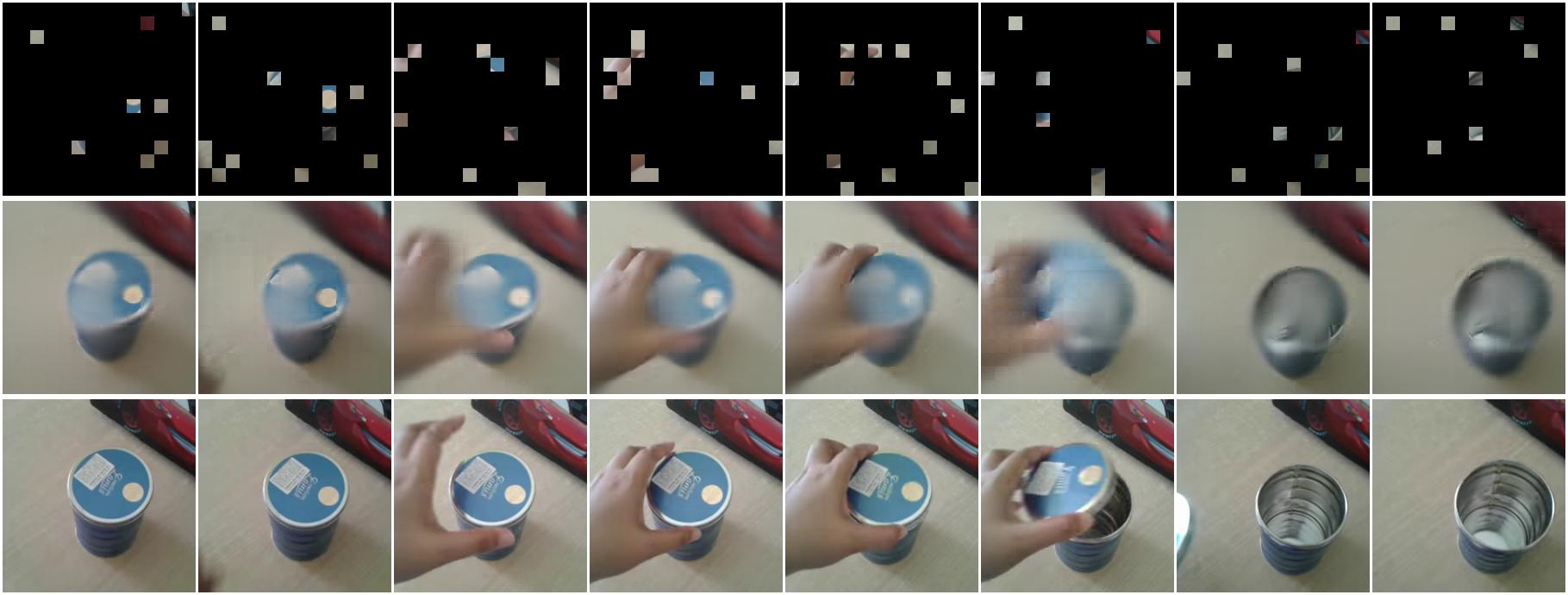} &
         \includegraphics[width=\percolwidth\linewidth, trim={0 0 0 0},clip]{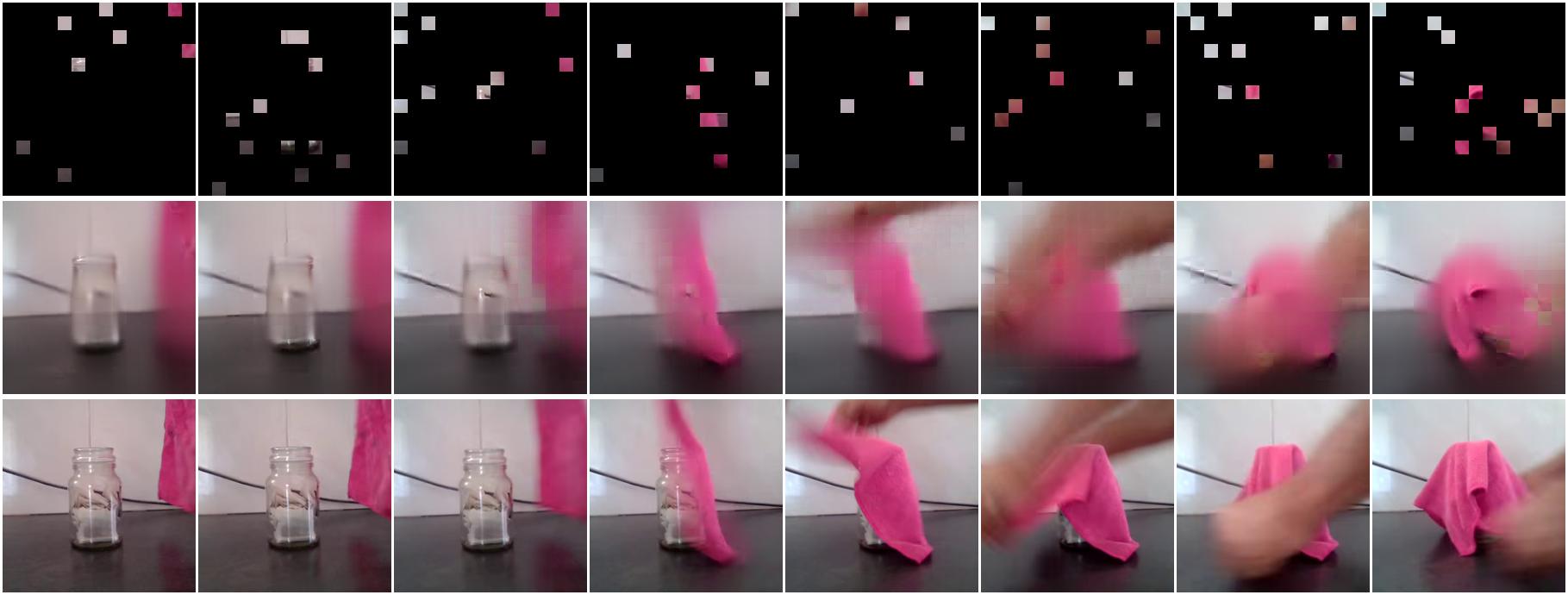} \\
         \includegraphics[width=\percolwidth\linewidth, trim={0 0 0 0},clip]{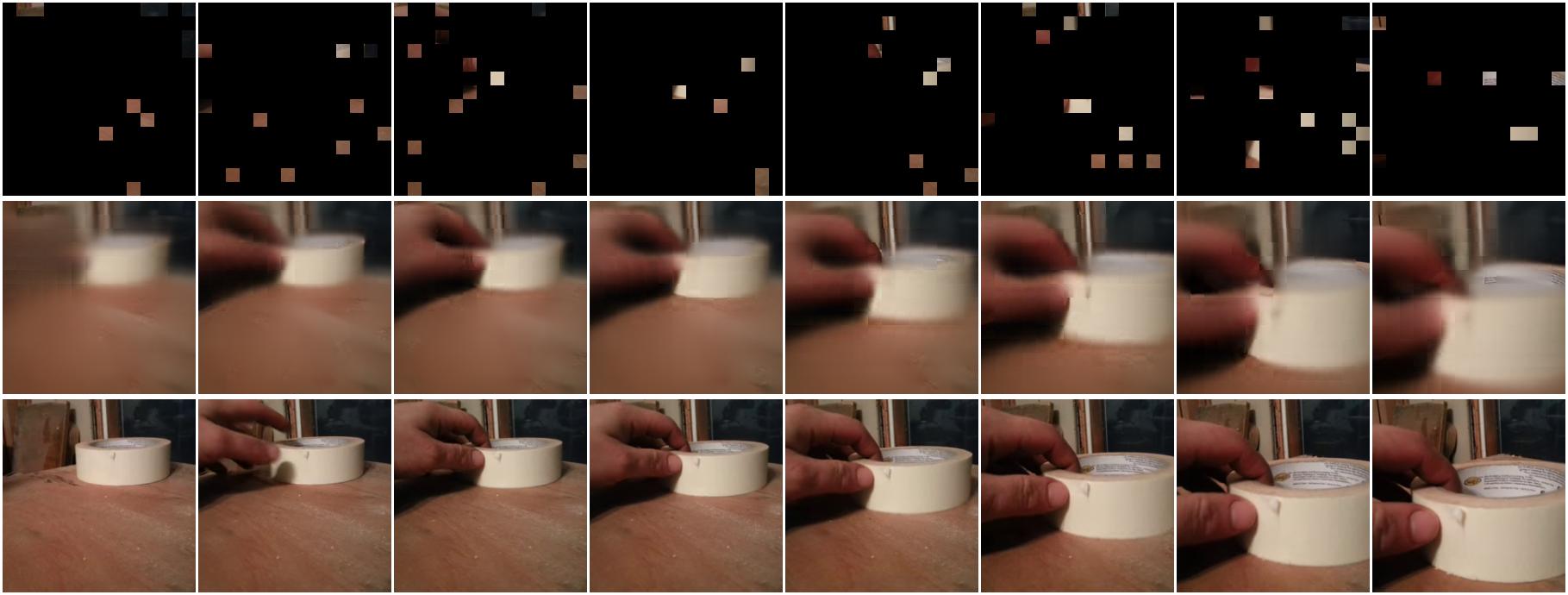} &
         \includegraphics[width=\percolwidth\linewidth, trim={0 0 0 0},clip]{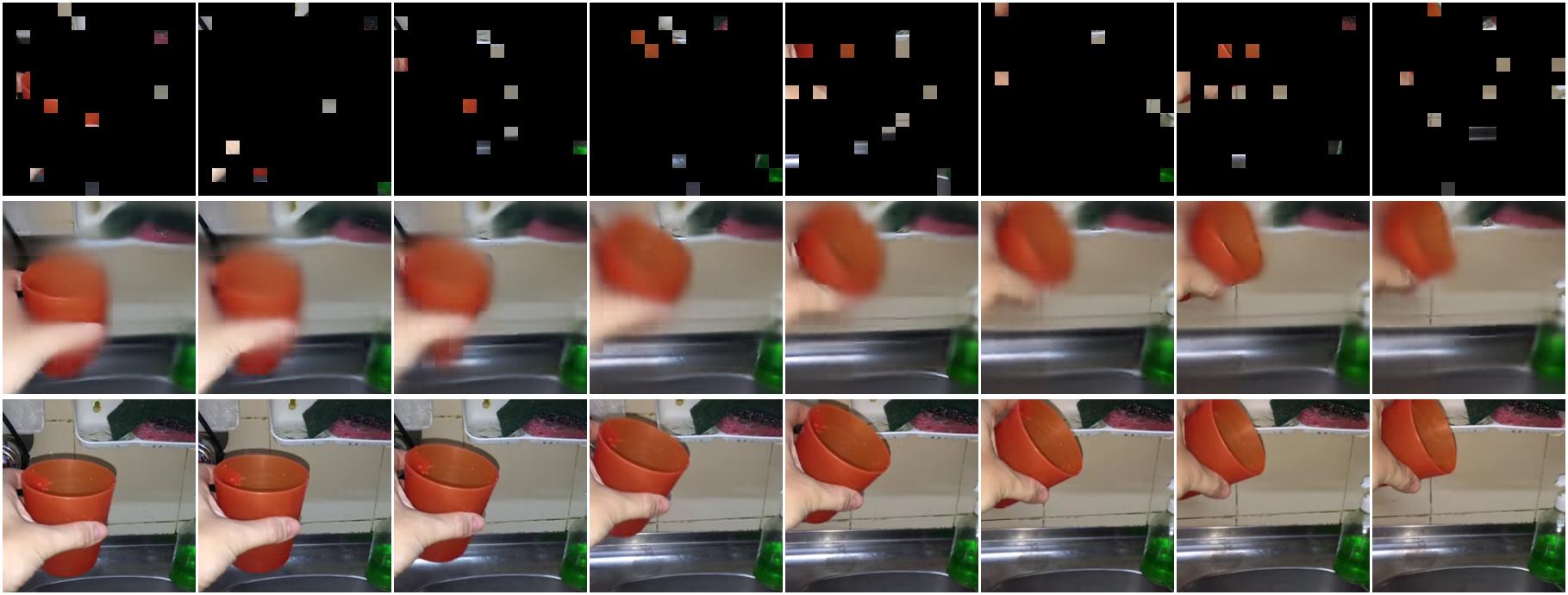} \\
        \multicolumn{2}{c}{ \color{VideoDark} \bf\kineticsShort}\\
         \includegraphics[width=\percolwidth\linewidth, trim={0 0 0 0},clip]{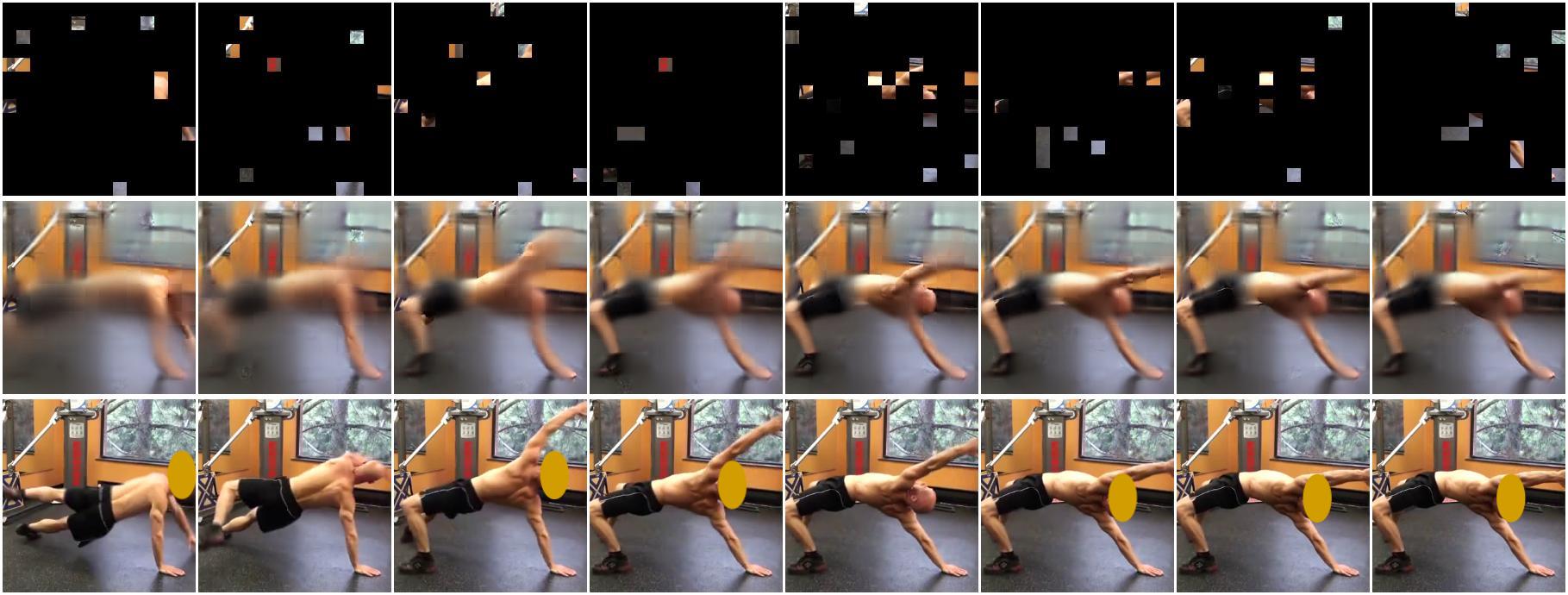} &
         \includegraphics[width=\percolwidth\linewidth, trim={0 0 0 0},clip]{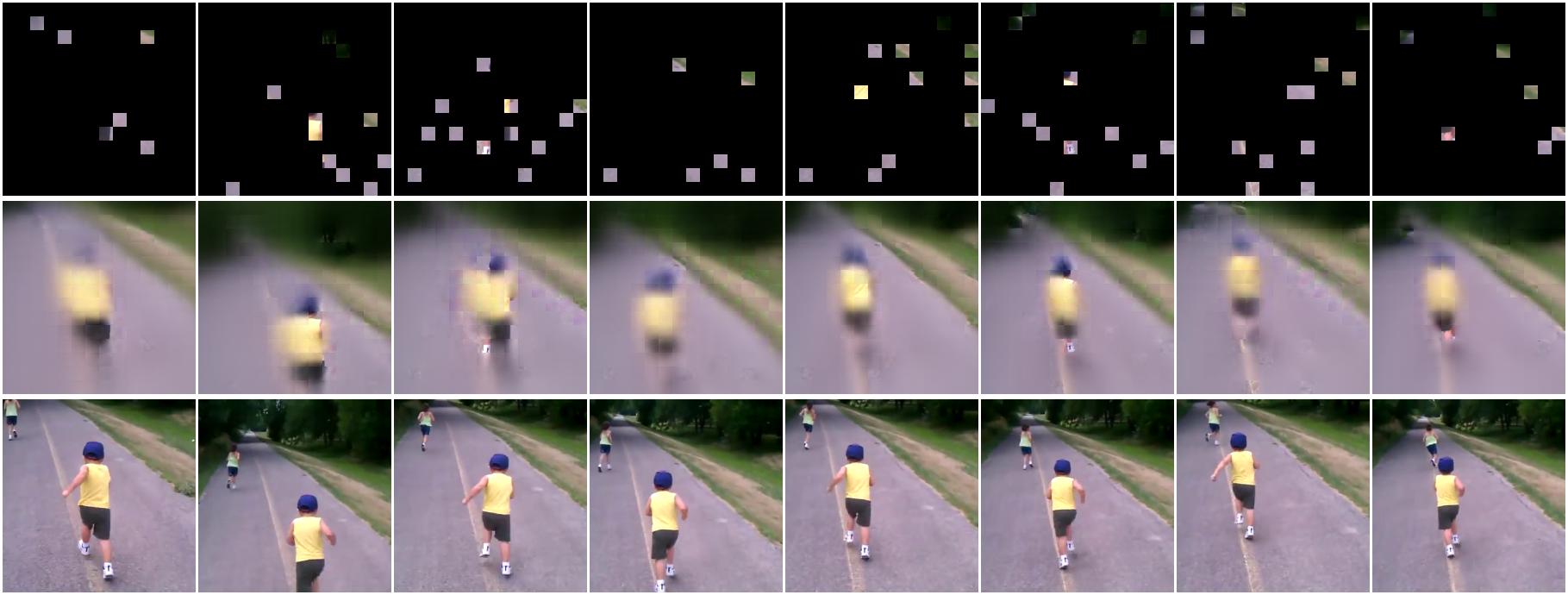} \\
         \includegraphics[width=\percolwidth\linewidth, trim={0 0 0 0},clip]{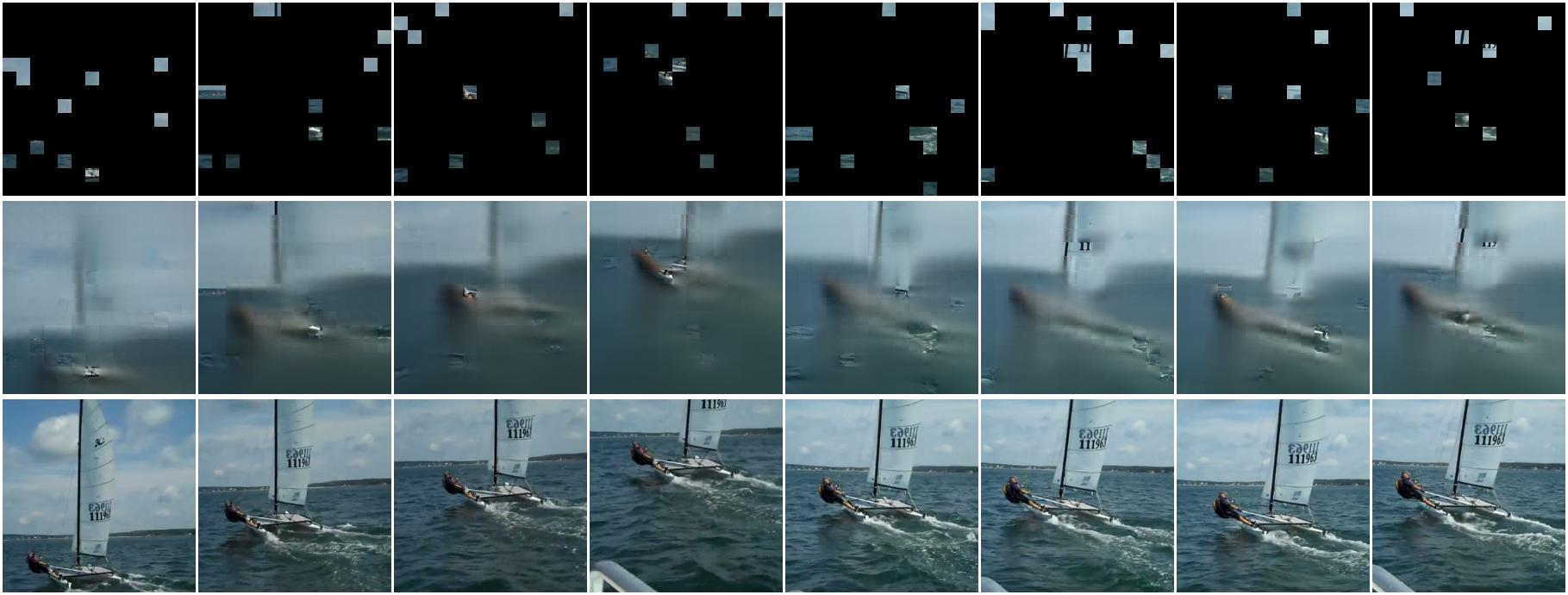} &
         \includegraphics[width=\percolwidth\linewidth, trim={0 0 0 0},clip]{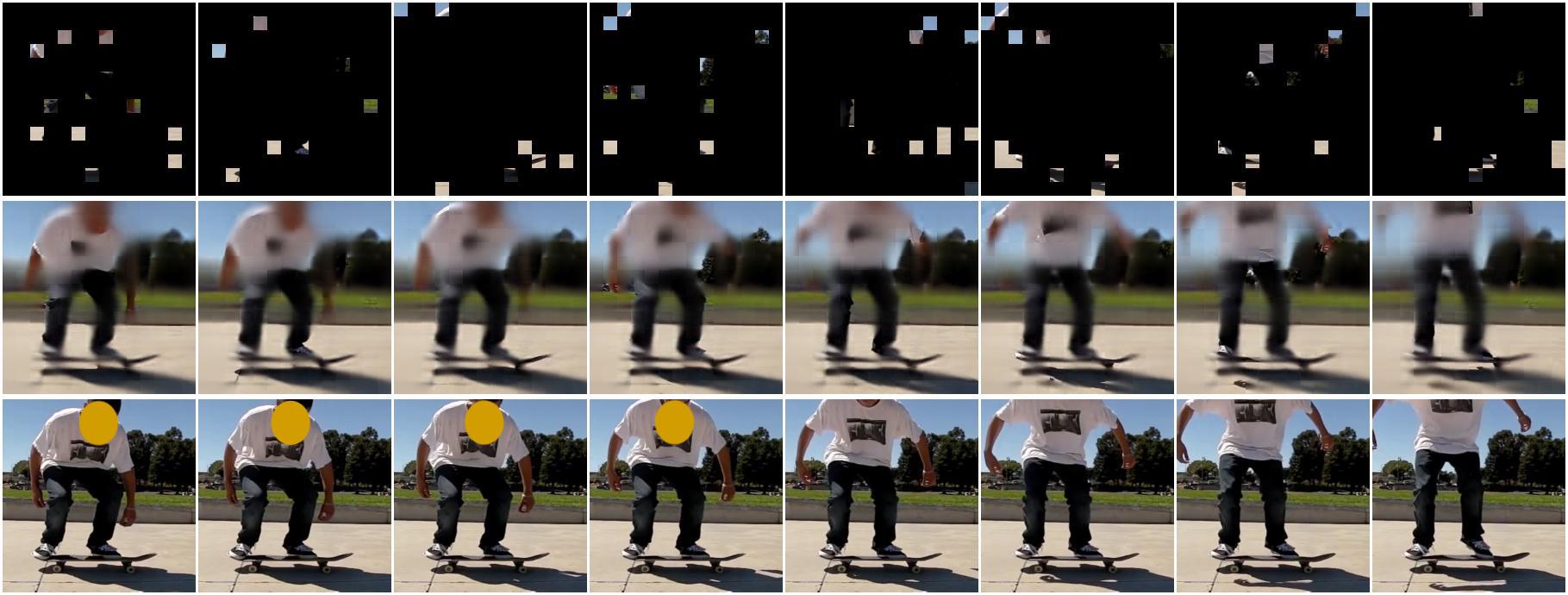} \\
        \multicolumn{2}{c}{\color{VideoDark} \bf\epicShort}\\
         \includegraphics[width=\percolwidth\linewidth, trim={0 0 0 0},clip]{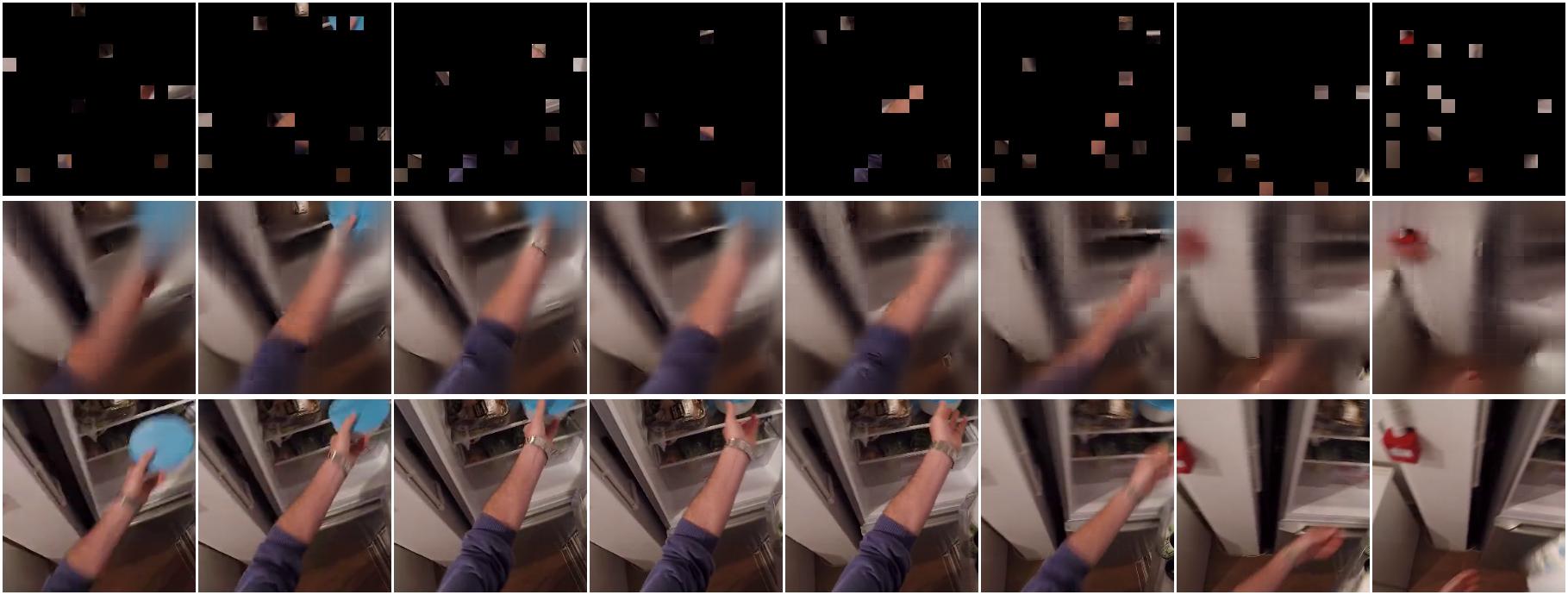} &
         \includegraphics[width=\percolwidth\linewidth, trim={0 0 0 0},clip]{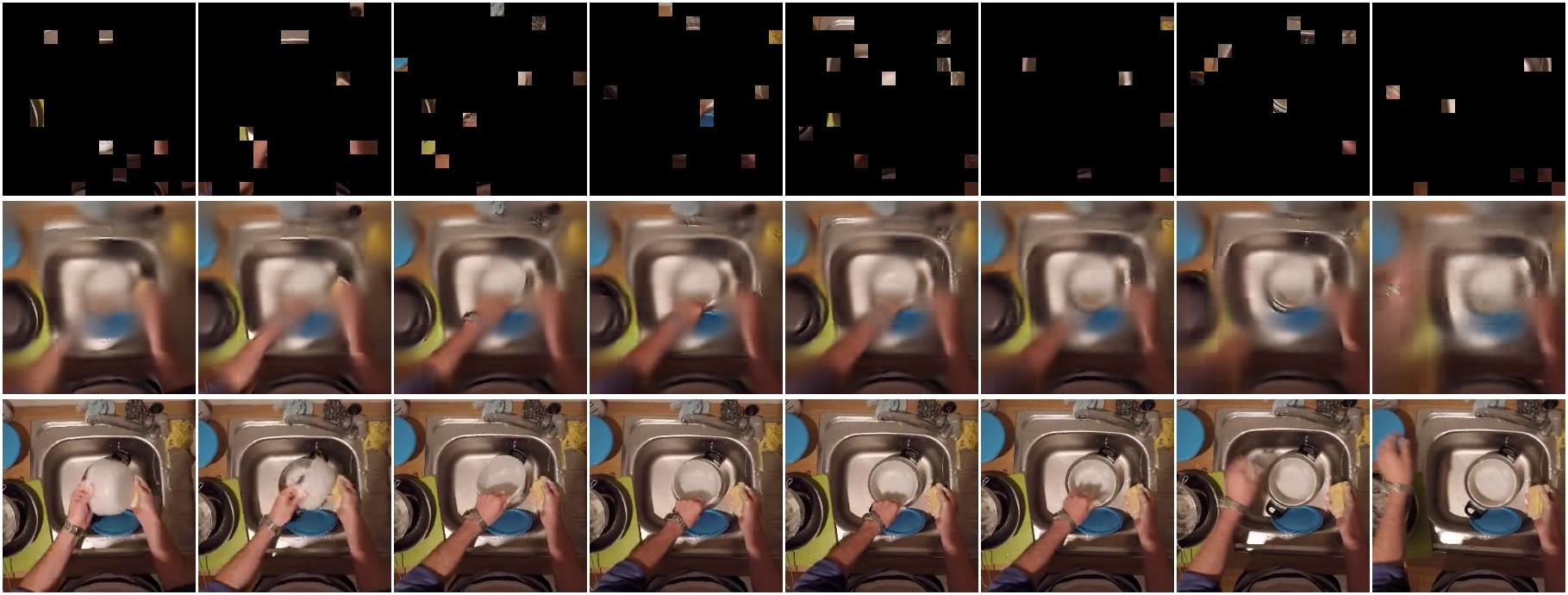} \\
         \includegraphics[width=\percolwidth\linewidth, trim={0 0 0 0},clip]{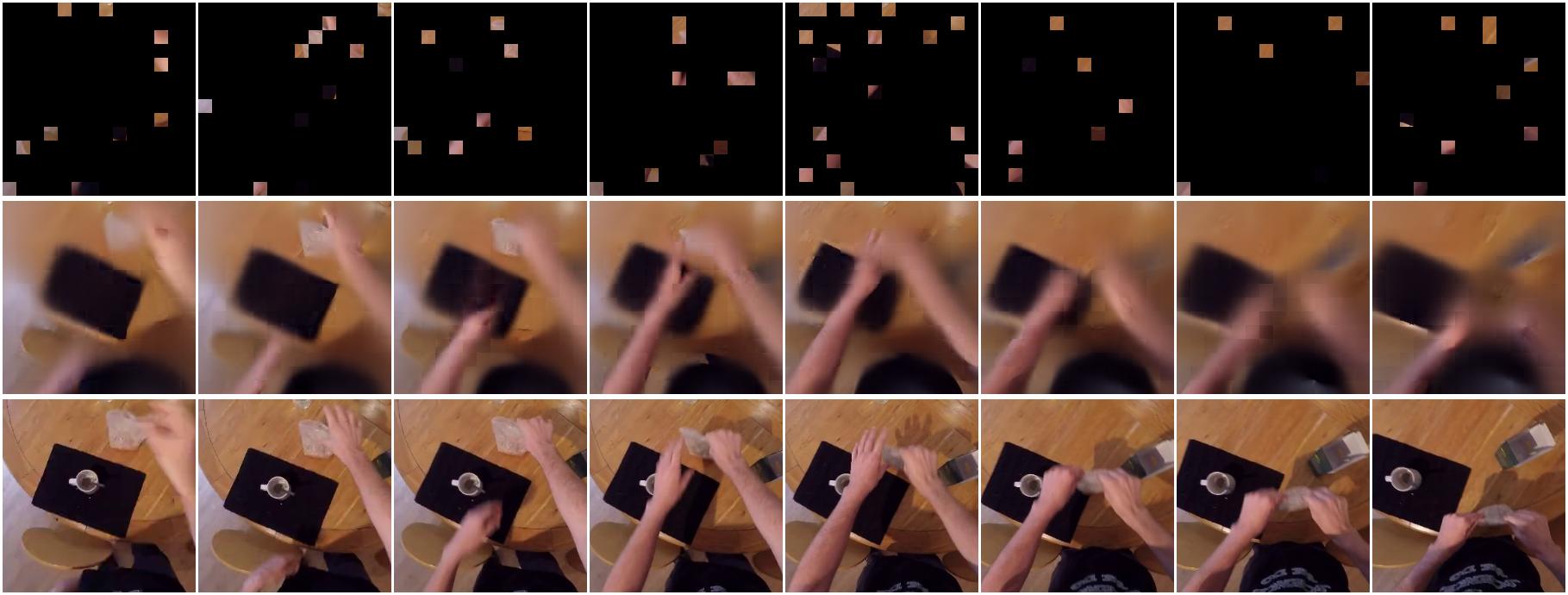} &
         \includegraphics[width=\percolwidth\linewidth, trim={0 0 0 0},clip]{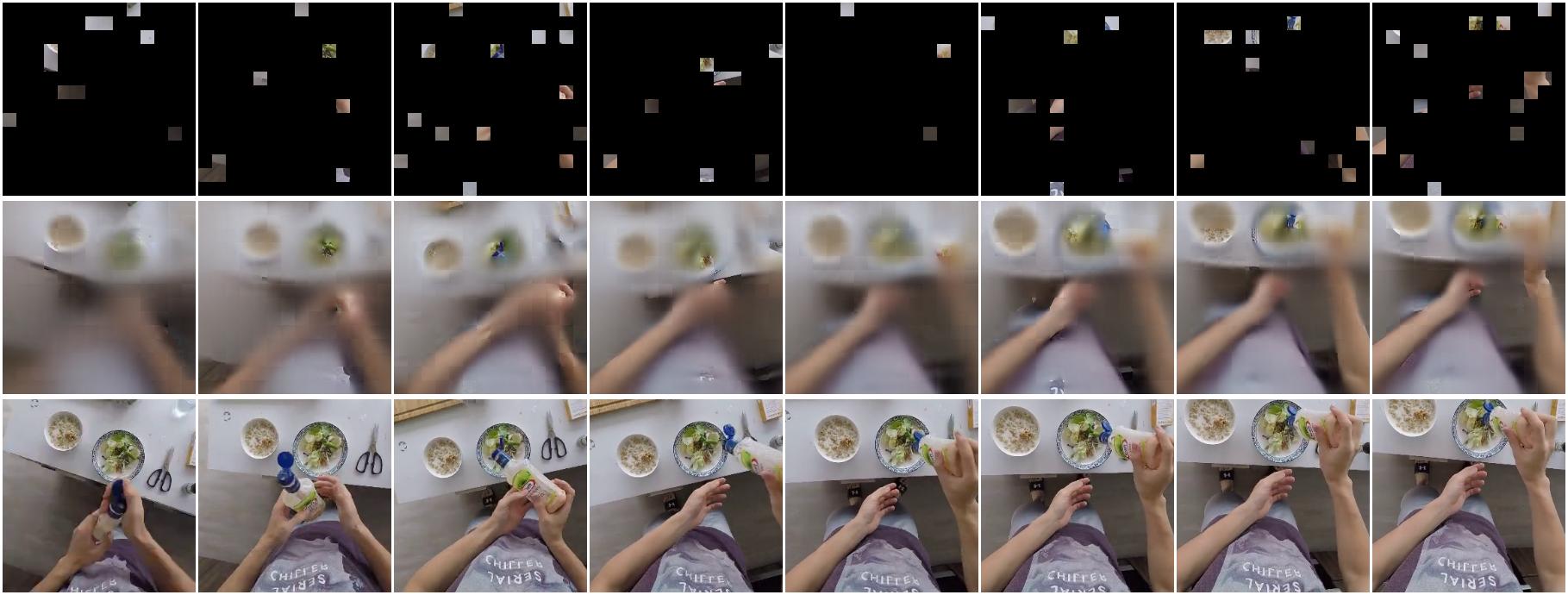} \\
     \end{tabular}}
     \captionof{figure}{{\bf Additional Reconstruction visualizations} using \OURS on different {\color{VideoDark}video} datasets. We show the model predictions for a masking ratio of 95\%. 
     }\label{fig:add_visualizations_video}
\end{table}
\begin{table}[t]
     \centering
     \setlength{\tabcolsep}{2pt}
     \newcommand{\raiseboxht}[0]{-0.45}
     \newcommand{\raiseboxlastht}[0]{\raiseboxht}
     \newcommand{\vizcropht}[0]{8.1cm}
     \newcommand{\vizcroplastht}[0]{16.3cm}
     \newcommand{\percolwidth}[0]{0.48}
     \newcommand{\perimgcolwidth}[0]{0.06}  %
     \hspace{-0.15in}  %
     \resizebox{\linewidth}{!}{
     \begin{tabular}{ccccccccccccccc}
         \includegraphics[width=\perimgcolwidth\linewidth, trim={0 0 0 0},clip]{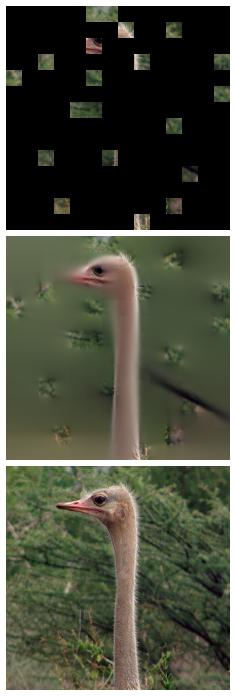} &
         \includegraphics[width=\perimgcolwidth\linewidth, trim={0 0 0 0},clip]{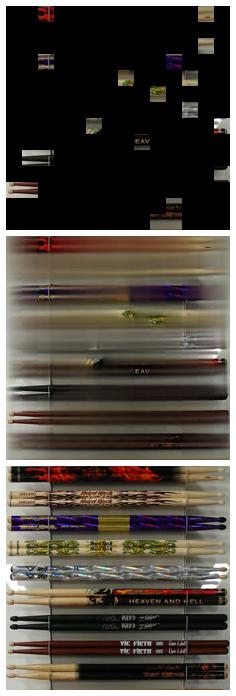} &
         \includegraphics[width=\perimgcolwidth\linewidth, trim={0 0 0 0},clip]{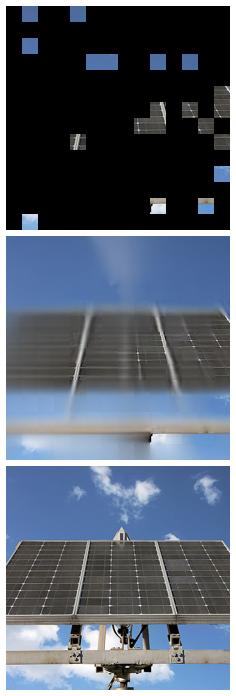} &
         \includegraphics[width=\perimgcolwidth\linewidth, trim={0 0 0 0},clip]{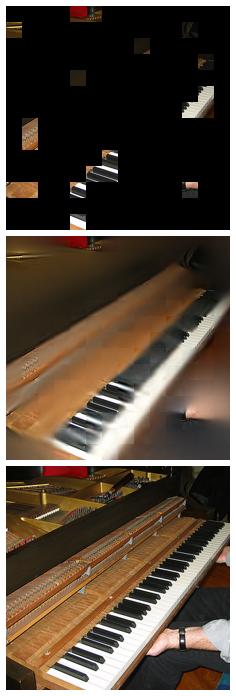} &
         \includegraphics[width=\perimgcolwidth\linewidth, trim={0 0 0 0},clip]{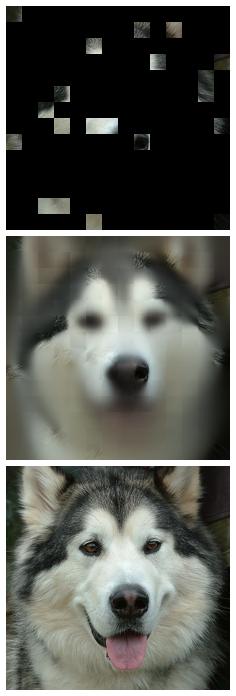} &
         \includegraphics[width=\perimgcolwidth\linewidth, trim={0 0 0 0},clip]{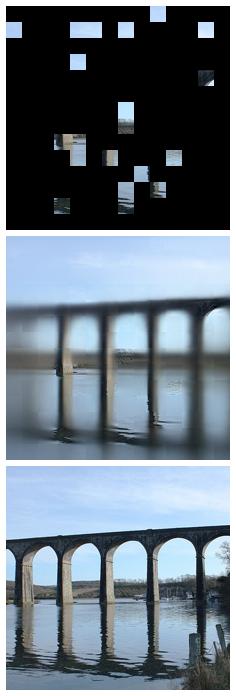} &
         \includegraphics[width=\perimgcolwidth\linewidth, trim={0 0 0 0},clip]{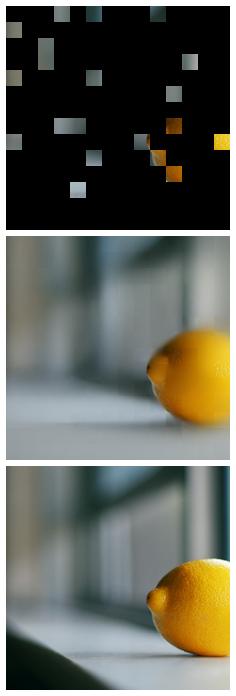} &
         \includegraphics[width=\perimgcolwidth\linewidth, trim={0 0 0 0},clip]{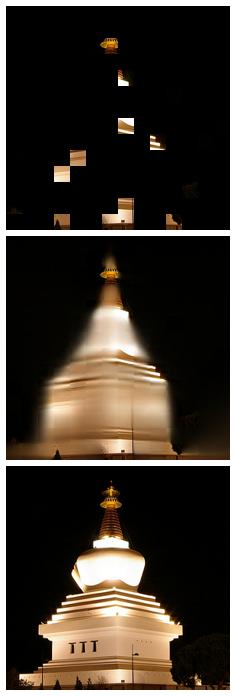} &
         \includegraphics[width=\perimgcolwidth\linewidth, trim={0 0 0 0},clip]{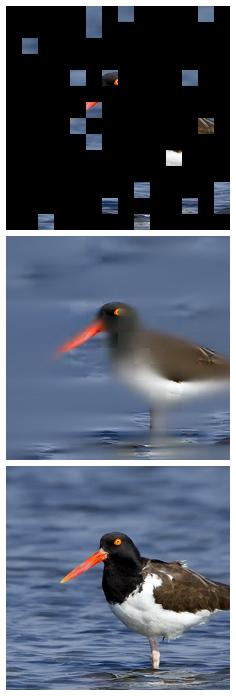} &
         \includegraphics[width=\perimgcolwidth\linewidth, trim={0 0 0 0},clip]{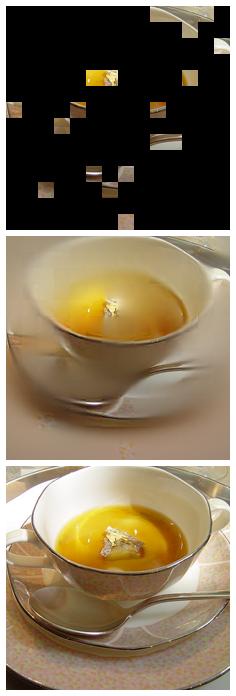} &
         \includegraphics[width=\perimgcolwidth\linewidth, trim={0 0 0 0},clip]{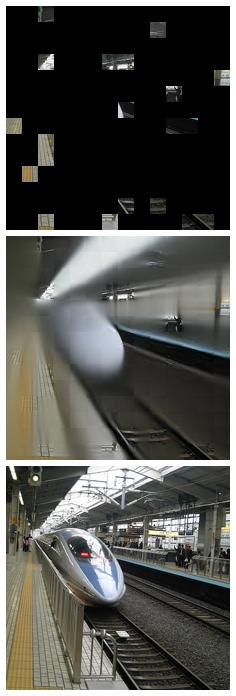} &
         \includegraphics[width=\perimgcolwidth\linewidth, trim={0 0 0 0},clip]{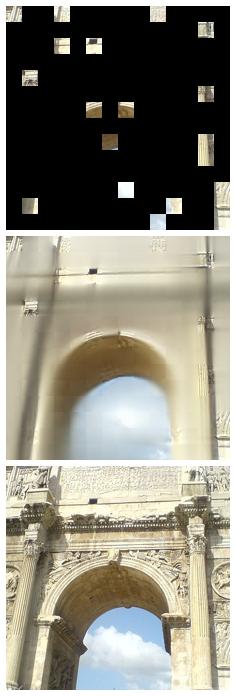} &
         \includegraphics[width=\perimgcolwidth\linewidth, trim={0 0 0 0},clip]{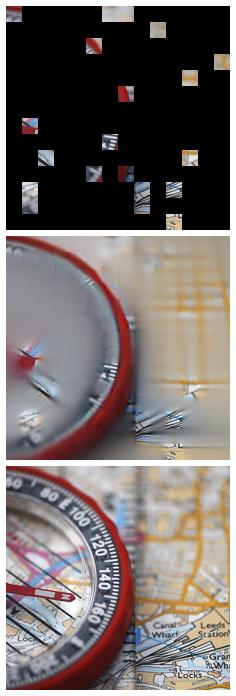} &
         \includegraphics[width=\perimgcolwidth\linewidth, trim={0 0 0 0},clip]{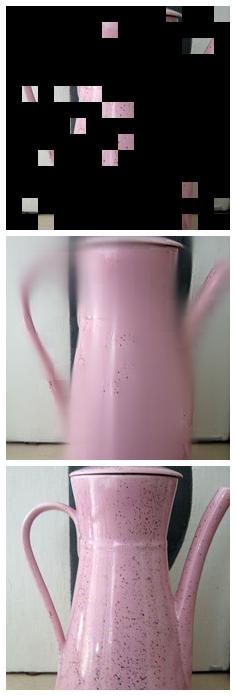} \\
         
     \end{tabular}}
     \captionof{figure}{{\bf Additional Reconstruction visualizations} using \OURS on the {\color{ImageDark} \imnetShort} image dataset. We show the model predictions for a masking ratio of 90\%.
     }\label{fig:add_visualizations_img}
\end{table}

\begin{table}[t]
    \centering
    \resizebox{\linewidth}{!}{
    \begin{tabu}{ll|cc}
        {\bf Setting} & {\bf Data} & {\bf \color{ImageDark} \imnetShort} & {\bf \color{VideoDark} \sthsthShort} \\
        \hline
        \OURS & \imnetShort + \sthsthShort frames & 82.7 & 64.2 \\
        \OURS & \imnetShort + \sthsthShort & {\bf 82.8} & {\bf 69.0} \\
        \rowfont{\color{DarkGray!70}} \mae (\cf~\cref{fig:compare_all_maes})	& \imnetShort & 83.4 & 59.5 \\
    \end{tabu}}
    \caption{{\bf Effect of extra data.}
    We train \OURS with the exact set of \imnetShort images and \sthsthShort frames  used during original \OURS pretraining with \imnetShort images and \sthsthShort videos. This follows our setup in~\cref{sec:ablations}, where we train \vitB for 800 epochs. While the \imnetShort image classification performance in both settings is comparable, the \sthsthShort video classification performance drops significantly by almost 5\% when trained only using frames and not video clips, although it is better than just training with \imnetShort images. This shows that the performance gains with \OURS are not merely due to the additional data being used for training.
    }\label{tab:extra_data}
\end{table}

\section{Additional Ablations}\label{sec:ablations_appdx}

{\noindent \bf Effect of extra data.}
Since \OURS is trained with extra data compared to \mae, one concern is whether the gains can be attributed to joint training or simply the extra frames. To that end, we experiment with training \mae with individual frames from \sthsthShort instead of videos.
To ensure an exact apples-to-apples comparison, we use the exact setup for \OURS, and simply convert each video input into individual frames, hence ensuring the exact same epochs, number of parameter updates, data, learning rates schedule etc. As we see in~\cref{tab:extra_data}, the \sthsthShort video classification performance drops by almost 5\% when trained with frames and not video clips, although it is better than training with only \imnetShort images. This ensures that the gains are indeed from jointly training on the two modalities, rather than simply using more data during training.

\end{document}